\documentclass[10pt, a4paper, onecolumn]{article}

\usepackage{arxiv}
\usepackage[toc]{appendix}
\usepackage[utf8]{inputenc} 
\usepackage[T1]{fontenc}    
\usepackage{hyperref}       
\usepackage{url}            
\usepackage{booktabs}       
\usepackage{amsfonts}       
\usepackage{nicefrac}       
\usepackage{microtype}      
\usepackage{lipsum}
\usepackage{amsmath}
\usepackage{cite}
\usepackage{amssymb}
\usepackage{verbatim}
\usepackage{enumerate}
\usepackage{mdwlist} 
\usepackage{color}

\usepackage{algpseudocode}
\usepackage{algorithmicx}
\usepackage{lineno}
\usepackage{framed,multirow}
\usepackage[final]{pdfpages}
\usepackage{latexsym}
\usepackage{wrapfig}

\usepackage{url}
\usepackage[table]{xcolor}
\usepackage{xcolor}
\usepackage{graphicx}
\usepackage{subfigure}
\usepackage{verbatim}
\usepackage{bm}
\usepackage{authblk}
\usepackage{tikz}
\usepackage{amsmath}
\usepackage[linesnumbered,ruled]{algorithm2e}
\usepackage[space]{grffile}
\usetikzlibrary{shapes,arrows}

\newcommand{\etal}{\textit{et al.}}

\title{A Comparative Survey of Vision Transformers for Feature Extraction in Texture Analysis
}

\author[1]{Leonardo Scabini}
\author[2]{Andre Sacilotti}
\author[1]{Kallil M. Zielinski}
\author[3]{Lucas C. Ribas}
\author[4]{Bernard De Baets}
\author[1,2]{Odemir M. Bruno}

\affil[1]{\small{S\~{a}o Carlos Institute of Physics, University of S\~{a}o Paulo, postal code 13560-970, São Carlos - SP, Brazil}}
\affil[2]{\small{Institute of Mathematics and Computer Sciences, University of S\~{a}o Paulo, postal code 135666-590, São Carlos - SP, Brazil}}
\affil[3]{\small{Institute of Biosciences, Humanities and Exact Sciences, São Paulo State University, postal code 15054-000, São José do Rio Preto, Brazil}}
\affil[4]{\small{KERMIT, Department of Data Analysis and Mathematical Modelling, Ghent University, Coupure links 653, postal code 9000, Ghent, Belgium}}

\begin{document}
\maketitle
\begin{abstract}
Texture, a significant visual attribute in images, has been extensively investigated across various image recognition applications. Convolutional Neural Networks (CNNs), which have been successful in many computer vision tasks, are currently among the best texture analysis approaches. On the other hand, Vision Transformers (ViTs) have been surpassing the performance of CNNs on tasks such as object recognition, causing a paradigm shift in the field. However, ViTs have so far not been scrutinized for texture recognition, hindering a proper appreciation of their potential in this specific setting. For this reason, this work explores various pre-trained ViT architectures when transferred to tasks that rely on textures. We review 21 different ViT variants and perform an extensive evaluation and comparison with CNNs and hand-engineered models on several tasks, such as assessing robustness to changes in texture rotation, scale, and illumination, and distinguishing color textures, material textures, and texture attributes. The goal is to understand the potential and differences among these models when directly applied to texture recognition, using pre-trained ViTs primarily for feature extraction and employing linear classifiers for evaluation. We also evaluate their efficiency, which is one of the main drawbacks in contrast to other methods. Our results show that ViTs generally outperform both CNNs and hand-engineered models, especially when using stronger pre-training and tasks involving in-the-wild textures (images from the internet). We highlight the following promising models: ViT-B with DINO pre-training, BeiTv2, and the Swin architecture, as well as the EfficientFormer as a low-cost alternative. In terms of efficiency, although having a higher number of GFLOPs and parameters, ViT-B and BeiT(v2) can achieve a lower feature extraction time on GPUs compared to ResNet50.
\end{abstract}

\section{Introduction}
\begin{figure*}[!htb]
    \centering
    \includegraphics[width=0.95\linewidth]{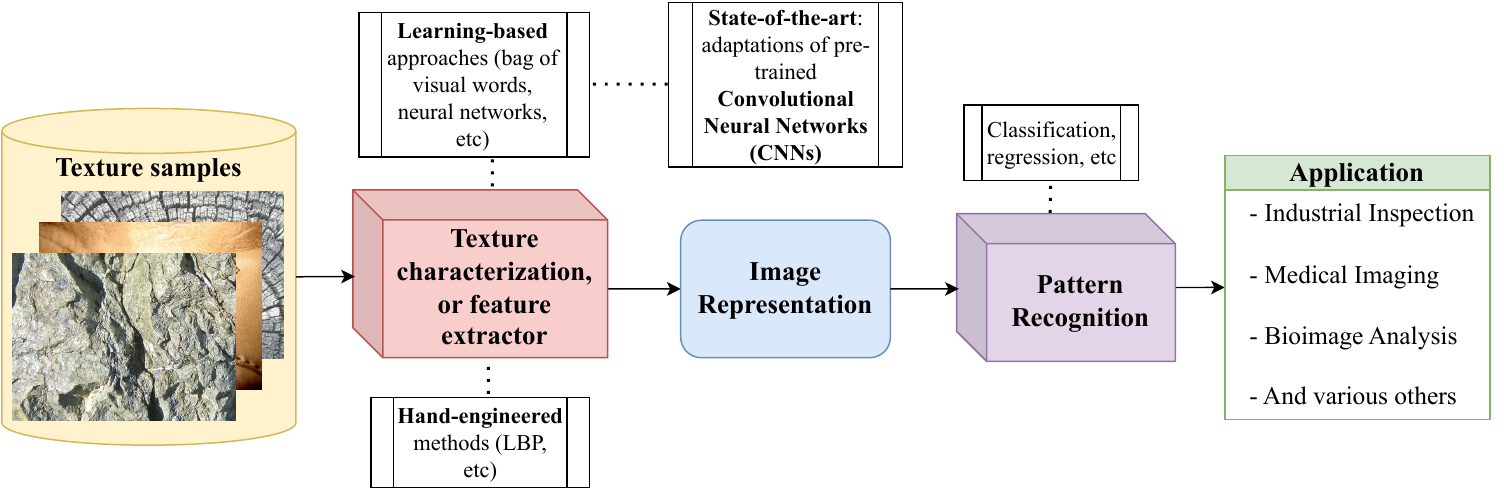}
    
  \caption{The usual pipeline in texture analysis. Texture samples must be encoded in meaningful image representations using feature extraction techniques, which could be either hand-engineered (usually designed specifically for textures), or based on learning models (e.g., from pre-trained deep neural networks). These representations can be used for pattern recognition tasks in a variety of applications.  
  \label{fig:intro}}
\end{figure*}

Computer Vision (CV) has become an extensive subfield of Artificial Intelligence, especially after the
proliferation of deep learning over the past decade. One of the subfields of CV, which dates back to the 60s, is texture analysis. For digital images, one abstract definition is that texture elements emerge from the local intensity constancy and/or variations of pixels producing spatial patterns roughly independently at different scales~\cite{scabini2020spatio}. Although general vision involves the combination of several other aspects such as shape and depth, texture alone is a fundamental characteristic that can suffice to solve many problems. Therefore, over the past decades several texture analysis methods have been 
proposed~\cite{liu2019bow,humeau2019texture}. These works led to many applications in industrial inspection~\cite{pietikainen1996texture}, medical imaging~\cite{kassner2010texture}, to name but a few. Figure~\ref{fig:intro} illustrates the usual texture analysis approach: a model extracts relevant information from a texture image to compose an image representation, which is used for pattern recognition tasks that rely on these textures. 

Deep-learning-based models for general vision tasks have been advancing rapidly. This revolution in the field started with Convolutional Neural Networks (CNNs)~\cite{lecun1989backpropagation,lecun1998gradient,krizhevsky2017imagenet}, a powerful neural architecture that still dominates many CV areas. However, CNNs may fail to achieve state-of-the-art (SOTA) performance on texture recognition tasks in comparison to 
hand-engineered approaches~\cite{scabini2019multilayer,scabini2020spatio}. More recently, Vision Transformers (ViTs) have started to dominate the CV literature while challenging CNNs, especially on image classification tasks~\cite{dosovitskiy2020image,touvron2022deit}. Nevertheless, little is known regarding the applicability of ViTs to texture analysis. 
This is even more concerning given the fact that this architecture
does not exhibit the common spatial inductive biases that CNNs have,
therefore lacking translation equivariance and locality awareness. 
To overcome these limitations, subsequent works have proposed strong pre-training of ViTs on large datasets and spatial embeddings of the input image.
Although these approaches usually improve the general performance, they do not ensure spatial invariance, which raises more questions about the performance of ViTs 
for texture recognition. Moreover, texture datasets and texture recognition applications usually have limited training data available, thus increasing the challenge for data-hungry models such as ViTs.

In this paper, we explore a wide range of ViTs when applied to texture analysis through transfer learning. Twenty-one ViT variants are selected in terms of architectural design and pre-training approach, including different paradigms (supervised or self-supervised) and different datasets (ImageNet-1k or 21k). Our main goal is to explore the potential and differences among so-called foundation models, which are known to perform well on general vision tasks, when directly applied to texture recognition. We also compare them with CNN and hand-engineered baselines applied to the same tasks.
The foundation models are employed by removing their classification head, freezing the pre-trained weights, and using them only to extract features. Linear classifiers are then trained on these features aiming at a variety of texture recognition tasks. These tasks encompass different scenarios such as measuring the robustness to changes in texture rotation, scale, and illumination, as well as discriminating color texture, material textures, and texture attributes.

\section{Background}

\subsection{Vision Transformers}
The transformer architecture~\cite{vaswani2017attention} is an effective deep learning mechanism for machine translation tasks with a more parallelizable architecture. The first architecture was designed as a stack of encoders and decoders, containing two main structures: Multihead Self-Attention (MSA) and a Feed Forward Network. First of all, consider a set of tokens and their embeddings (e.g., words and word embeddings) combined into a matrix $X$. The first step is to transform these inputs by projecting them using linear layers, obtaining a query matrix $Q_i=XW_{Q_i}$, where $W_{Q_i}$ represents the query weights, and matrices $K_i=XW_{K_i}$ and $V_i=XW_{V_i}$ representing the keys and values, respectively, and their corresponding weights. The self-attention mechanism for all tokens is obtained by:
\begin{equation}
\label{eq:self-attention}
\text{Attention}(Q_i, K_i, V_i) = \text{softmax}(\frac{Q K^{T}}{\sqrt{d}}) V\,,
\end{equation}
where the softmax function is taken over the horizontal axis, and $d$ is the hidden dimension of the model (embedding size). A single self-attention mechanism is referred to as an attention head and MSA is achieved by stacking $s$ attention heads in parallel, 
each with individual trainable weights:
\begin{equation}
\label{eq:MSA}
\text{MSA}(Q, K, V) = [\text{Attention}(Q_1, K_1, V_1);\ldots;\text{Attention}(Q_s, K_s, V_s)]W_O\,,
\end{equation}
where $[;]$ denotes the concatenation of the self-attentions, and $W_O$ represents the weights for a final linear projection after the concatenation. A Feed Forward Network is applied over the output of the MSA, which is a simple MLP with two layers. Additionally, layer normalization and residual connections are added between the layers, finally composing a transformer block. A standard transformer network is then the combination of a series of transformer blocks, followed by an output layer depending on the task at hand.

\begin{figure*}[!htb]
    \centering
    \subfigure[General structure of a Vision Transformer and some of the different options that can be selected at each stage.]{
    \includegraphics[width=0.95\linewidth]{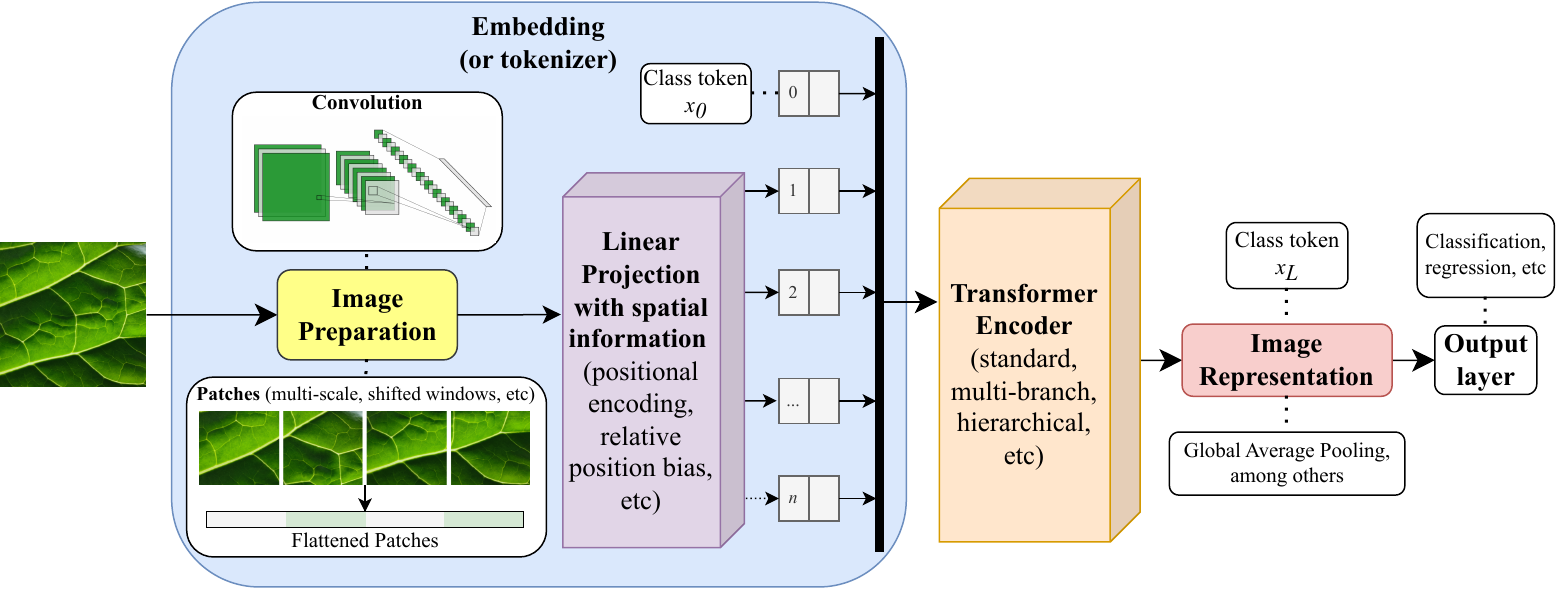}}
    \\
    \subfigure[The most common type of Transformer Encoder (adapted from~\cite{dosovitskiy2020image}).]{\includegraphics[width=0.75\linewidth]{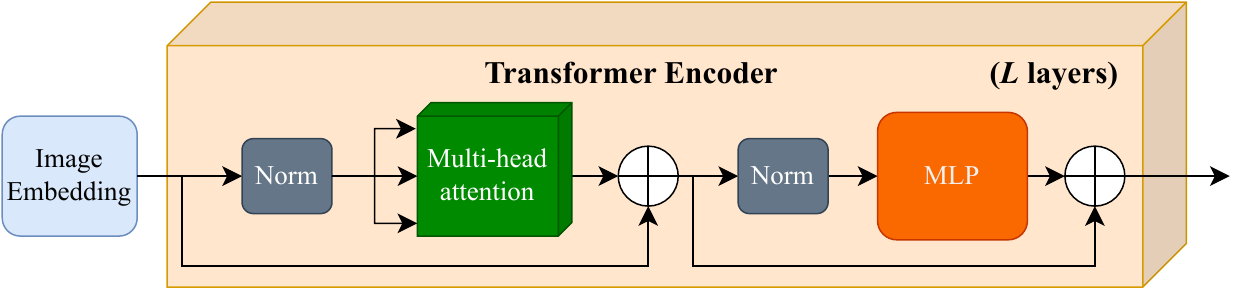}}
    
  \caption{The general elements of a Vision Transformer (a). One of its most important modules is the image embedding (a.k.a.\ tokenizer), which is responsible for preparing the pixels in a way that Transformer Encoders (b) can learn and extract meaningful visual patterns. \label{fig:ViT}}
\end{figure*}

 More recently, Vision Transformers (ViTs)~\cite{dosovitskiy2020image,touvron2022deit} have been dominating the CV literature challenging CNNs. They have been applied to a lot of different visual tasks, including, but not limited to: image classification~\cite{dosovitskiy2020image,touvron2022deit,beit,beitv2}, 
 object detection~ \cite{NEURIPS2021_dc912a25,Hong_2022_CVPR}, image segmentation~\cite{9746172}, and 
 super-resolution~\cite{https://doi.org/10.48550/arxiv.2108.11084}. Results demonstrate that ViTs achieve SOTA performance on CV tasks, on par with CNNs. Figure~\ref{fig:ViT} (a) shows the general structure of a ViT, and some of the main architectural choices in recent works.

 ViTs offer an efficient way to adapt the transformer architecture to images by representing them as a sequence of 2D patches. Consider an RGB image $I \in \mathbb{R}^{w \times h \times 3}$, with width $w$ and height $h$, fed into a ViT backbone (as in \cite{dosovitskiy2020image}) $B= (T_1,\ldots,T_L)$, consisting of $L$ sequential transformer blocks. The image $I$ is firstly reshaped into a sequence of flattened 2D patches $I_p \in \mathbb{R}^{n \times (3p^2)}$, where 3 represents the RGB colorspace, $p \times p$ is the patch size, and $n=\frac{wh}{p^2}$ represents the number of patches (input sequence length). $I_p$ is then projected using a trainable linear projection (linear layer) $E \in \mathbb{R}^{(3p^2) \times d}$, where $d$ is the constant hidden size of the transformer architecture. The patch embeddings $I_{e}$ are obtained by adding (element-wise sum) a positional encoding layer $E_{pos} \in \mathbb{R}^{(n+1) \times d}$, including spatial inductive bias, into the transformer. $I_{e}$ is then fed into a sequence of $L$ transformer encoders (such as shown in Figure~\ref{fig:ViT} (b)), which can be trained by adding an MLP classifier at the end. This is possible since the ViT includes a learnable class embedding (or class token):
 \begin{equation}\label{eq:classtoken}
   x^0_l \in \mathbb{R}^d,  
 \end{equation}
among the embeddings of each layer $l$, which encodes the information in a one-dimensional vector with $d$ features. Therefore, the class embedding at the last transformer layer, $x^0_L$, serves as an image representation where a classification head can be attached. In this work, we focus on this embedding as an off-the-shelf image representation for texture recognition tasks using simpler/linear classifiers, so the common MLP head of ViTs is not considered.

\subsection{Texture Analysis}

Texture analysis is a subfield of CV with roots in the 1960s. Although being an abstract concept, with no widely accepted formal definition, texture refers to the perceived surface properties or structure of objects, which may include roughness, smoothness, coarseness, or fineness. It can also be seen as a pattern of local variations in color and brightness. The human visual system is adept at recognizing and distinguishing textures, allowing us to differentiate between many things in our environment. Therefore, recognizing textures in digital images is critical to solve many CV problems and tasks such as feature extraction, classification, segmentation, synthesis, among others. As a result, a variety of methods has been developed over the past decades~\cite{liu2019bow,humeau2019texture}, paving the way for potential applications in fields such as industrial inspection~\cite{pietikainen1996texture}, medical imaging~\cite{kassner2010texture}, among others.

For many years, the predominant approach to texture recognition was based on hand-engineered models or features to describe textures. Mathematical methods for the description of textural patterns usually consider properties such as statistics \cite{ojala2002multiresolution}, frequency \cite{hoang2005color}, complexity/fractality \cite{backes2012color,scabini2020spatio}, and others \cite{zhang2002brief}. Statistical methods investigate local measures based on grayscale co-occurrences, the most widely used being the Local Binary Patterns (LBP)~\cite{ojala2002multiresolution}, which have influenced various subsequent techniques. Another approach to texture analysis involves transforming the image into the frequency domain, where various methods such as Gabor filters~\cite{hoang2005color} 
have been proposed. 
Complexity approaches fall within the model-based paradigm, such as methods based on fractal 
dimension~\cite{backes2012color} or network science~\cite{cantero2018importance,scabini2020spatio}.

  
After the popularization of learning-based models for object recognition, many such methods were also specifically designed for tackling texture. For instance, there have been various studies involving deep CNNs for texture recognition by using transfer learning. The most common approach for the transfer-learning of foundation vision models is to fine-tune a pre-trained network for the desired task. However, even if these modes are pre-trained, it is impossible to predict how much fine-tuning data would be necessary to achieve satisfactory performance. In the case of texture analysis, previous works have shown~\cite{scabini2019multilayer} that fine-tuning or training these models from scratch may result in poor performance on texture recognition, particularly due to overfitting caused by a lack of large-scale texture datasets.

Some studies explore the transfer-learning of deep CNNs by using convolutional layers only for extraction of texture features, freezing their parameters, and using a dedicated classifier trained separately. This approach is also known as ``features-off-the-shelf", or deep convolutional activation features~\cite{donahue2014decaf}, and is a simple and fast way to transfer learning from foundation vision models. Cimpoi~\etal~\cite{cimpoi2014describing} proposed one of the first contributions on the subject, comparing the efficiency of different deep CNN architectures and approaches for feature extraction. Subsequently, many works have been proposed following these principles. One of the latest techniques, named Random encoding of Aggregated Deep Activation Maps (RADAM)~\cite{scabini2023radam}, performs multi-depth deep feature aggregation and trains randomized auto-encoders for each image to produce an encoded representation. This method does not fine-tune the CNN backbone, and results demonstrate that these locally learned representations provide SOTA performance on texture recognition. 

Another approach consists of end-to-end architectures that enable the training of new texture-specific modules/layers along with the fine-tuning of pre-trained CNN backbones. Zhang~\etal~\cite{DeepTEN} proposed an orderless encoding layer on top of a deep CNN,
called Deep Texture Encoding Network (Deep-TEN), which allows for images of arbitrary size. Yang~\etal~\cite{yang2022} proposed DFAEN (Double-order Knowledge Fusion and Attentional Encoding Network), which aggregates first- and second-order information for encoding texture features. Fine-tuning is needed in these methods to adapt the backbone to the new architecture along with a new classification layer since they contain new randomly initialized parameters.

\section{Transformer-based texture analysis}

As discussed above, SOTA texture recognition models~\cite{yang2022,scabini2023radam} adopt pre-trained CNN backbones for texture feature extraction, aggregation and encoding, achieving promising results. However, the fixed/limited size of a CNN's receptive field may struggle to model the correlations among global features and long-distance pixel relationships, which are critical for many tasks involving textures. On the other hand, the transformer architecture 
excels at capturing these patterns, which suggests they could be valuable alternatives. Nevertheless, since ViT architectures are quite recent, only a few studies have specifically focused on texture 
aspects. For instance, some works explore textures in transformers for image super-resolution~\cite{yang2020learning,yao2021depth} and remote sensing~\cite{yang2022msfusion}. For texture recognition, our focus in this work, a couple of works have analyzed ViTs in specific cases. In~\cite{zhang2021material}, the authors used ViTs for the recognition of steel texture blocks, showing that a custom transformer architecture can obtain a higher accuracy than a standard CNN and machine learning models. Another study analyzed the ViT architecture in the field of building and construction material recognition~\cite{soleymani2021construction}, demonstrating the capability to deal with imbalanced datasets, achieving a higher accuracy compared to classical CNNs. A more recent work~\cite{tao2023vitalnet} introduces a hybrid transformer model for the localization of anomalies on industrial textured surfaces.

Despite some efforts to adopt ViT models for texture analysis, there is a lack of understanding of how or why this architecture works for different types of texture, and the impacts and differences of the several ViT variants quickly emerging in the literature. No study has analyzed ViTs for texture recognition in general, considering well-known benchmarks, robustness issues, impacts of different architectural choices, pre-trainings, and so on. Prior research has focused on specific problems related to texture analysis, which is not sufficient to promote ViTs as the next SOTA in this area. Therefore, we focus on these aspects by proposing a comprehensive evaluation of a variety of known ViTs when applied to a wide range of texture recognition tasks.

\subsection{Selected ViTs}

The literature on ViTs has been quickly advancing in the past three years due to the success of one of its first implementations for image recognition~\cite{dosovitskiy2020image}. It would be impossible to cover here all the models proposed under the ViT umbrella
in this period. For this purpose, we select a set of different ViT variants considering the most prominent differences in their architecture, pre-training, and computational budget. Table~\ref{tab:taxonomy} shows the main properties of the selected variants. Additionally, we give a more detailed description of each ViT below:
\begin{table*}[!htb]
	\centering
	\caption{\label{tab:taxonomy}Taxonomy of the ViT variants used in this work and the baselines considered for comparison. We indicate the pre-training that was used by the variants employed in this work (we used ResNet50 and DeiT3 with both IN-1k and IN-21k versions), where $a \Rightarrow b$ means that the model was pre-trained on dataset $a$, fine-tuned on dataset $b$, and then used for feature extraction. The feature extraction costs consider a 224x224 RGB input, and the size $d$ indicates the dimensionality of the feature vector. For ViTs and ResNet50, the GFLOPs and the number of parameters refer only to the model backbone used for feature extraction, i.e., after removing the classification head. }
 \quad
	\resizebox{\linewidth}{!}{
	\begin{tabular}{c|c|cc|cccc}
		\hline
		&& \multicolumn{2}{c|}{pre-training} & \multicolumn{3}{c}{feature extraction cost}\\
    	model & embedding & dataset& paradigm &$d$&GFLOPs & param. (m)  \\
    	\hline
     hand-eng.\ baseline (LBP)~\cite{ojala2002multiresolution}& N.A. & N.A. & N.A. & $\approx 256$ & $\approx 0.05$ & 0 \\
     CNN baseline (ResNet50~\cite{he2016deep,beyer2022knowledge})& convolutional & ImageNet-1k and \textbf{21k}& supervised & 2048 & 4.1 & 25.5 \\
     \hline

    CoaT-Li-Mi~\cite{xu2021co}& convolutional & ImageNet-1k & supervised & 512 & 2.0 & 10.5 \\ 
    CoaT-Mi~\cite{xu2021co}& convolutional & ImageNet-1k & supervised & 216 & 7.2 & 10.1 \\ 
    MobileViT-S~\cite{mehta2021mobilevit}& convolutional & ImageNet-1k &  supervised & 640 & 1.4 & 4.9 \\

    MobileViTv2~\cite{mehta2022separable}& convolutional & ImageNet-1k & supervised  & 512 & 1.4 & 4.4 \\

    EfficientFormer-L1~\cite{li2022efficientformer}& patches & ImageNet-1k & supervised & 448 & 1.3 & 11.4 \\ 
         \hline
   EfficientFormer-L3~\cite{li2022efficientformer}& patches & ImageNet-1k & supervised & 512 & 3.9 & 30.4 \\ 

    ViT-B/16~\cite{dosovitskiy2020image}& patches  & \textbf{ImageNet-21k} & supervised & 768 & 16.9 & 85.8 \\
    ViT-B/16-DINO~\cite{caron2021emerging}& patches & ImageNet-1k & self-supervised & 768 & 16.9 & 85.8 \\
    ViT-B/16-SAM~\cite{chen2022vision}& patches & ImageNet-1k & supervised & 768 & 16.9 & 85.8 \\
    DeiT-B/16~\cite{touvron2021training}& patches & ImageNet-1k & supervised & 768 & 16.9 & 85.8 \\
    
    DeiT3-B/16~\cite{touvron2022deit}& patches & ImageNet-1k and \textbf{21k}& supervised & 768 & 16.9 & 85.8 \\

    CrossViT-B~\cite{chen2021crossvit}& patches & ImageNet-1k & supervised & 1152 & 20.1 & 103.9 \\

     ConViT~\cite{d2021convit}& convolutional & ImageNet-1k & supervised & 768 & 16.8 & 85.8 \\

    GC ViT-B~\cite{hatamizadeh2022global}& convolutional & ImageNet-1k & supervised & 1024 & 13.9 & 89.3 \\

   MViTv2-B~\cite{li2022mvitv2} & patches & ImageNet-1k & supervised & 768 & 8.9 & 50.7 \\


   CaiT-S24~\cite{touvron2021going}& patches & ImageNet-1k & supervised & 384 & 8.6 & 46.5 \\


    XCiT-M24/16~\cite{ali2021xcit}& patches & ImageNet-1k & self-supervised & 512 & 15.8 & 83.9 \\ 
    
    BeiT-B/16~\cite{beit}& patches & \textbf{ImageNet-21k} & self-supervised & 768 & 12.7 & 85.8 \\
   BeiTv2-B/16~\cite{beitv2}& patches & \textbf{ImageNet-1k $\Rightarrow$ 21k} & self-sup. $\Rightarrow$ sup. & 768 & 12.7 & 85.8 \\
    Swin-B~\cite{liu2021swin}& patches $+$ shifted windows & \textbf{ImageNet-21k} & supervised & 1024 & 15.1 & 86.7 \\ 

    	\hline
	\end{tabular}
 }
\end{table*}

\begin{itemize}

    \item \textbf{ViT-B/16}\cite{dosovitskiy2020image}: This was one of the first successful computer vision variants of the transformer model, proposed by Dosovitskiy et al. The model employed in this work, ViT-B/16, has a base size "B" encompassing  12 layers, a hidden dimension of size 768, and 12 attention heads. This configuration processes input images by dividing them into non-overlapping patches of $16\times16$ pixels, hence the B/16 designation. Each patch is then linearly embedded into a flat vector and passed through the transformer layers for further processing.

    \item \textbf{CoaT} \cite{xu2021co}: Co-scale conv-attentional image Transformers (CoaT) contain two mechanisms to improve ViTs on image classification: (i) the co-scale mechanism, maintaining separate encoder branches at different scales while allowing for attention across these scales. A serial and a parallel block were created to perform fine-to-coarse, coarse-to-fine, and cross-scale image modeling; (ii) a conv-attention module that incorporates convolutions in the factorized attention module for relative position embeddings, resulting in a considerably improved computational cost compared to traditional self-attention layers in transformers.
    The authors introduced two architectures: CoaT-Lite, which exclusively uses serial blocks to sequentially process down-sampled image features, and CoaT, which incorporates both serial and parallel blocks with the co-scale mechanism.
    Additionally, the authors test CoaT and CoaT-Lite across various model sizes: Tiny, Mini, Small, and Medium. Consequently, CoaT-Li-Mi and CoaT-Mi refer to the CoaT-Lite Mini and CoaT Mini variants, respectively.

    \item  \textbf{MobileViT-S}~\cite{mehta2021mobilevit}: The design of this architecture targets mobile vision applications, focusing on compactness, general purpose, and minimal latency. 
    For this purpose, the network integrates key characteristics from CNNs, such as spatial inductive biases and reduced sensitivity to data augmentation, with those from ViTS, including input-adaptive weighting and global processing capabilities~\cite{mehta2021mobilevit}.
    By incorporating properties from both CNNs and ViTs, the MobileViT achieves a discriminative representation using a low number of parameters and simple training approaches, such as basic augmentation techniques. 
    The MobileViT model has three size variations typically used in mobile applications: small, extra small, and extra extra small. In this work, we used the small version, which has 5.6 million parameters.

    \item \textbf{MobileViTv2}~\cite{mehta2022separable}: Although MobileViT models exhibit high performance and have few parameters compared to light-weight CNNs, they still face the issue of high latency, primarily due to the multi-headed self-attention. To overcome this limitation, MobileViTv2, an enhanced version of MobileViT, introduces a separable self-attention mechanism with linear complexity that calculates context scores relative to a latent token.

    \item  \textbf{EfficientFormer}~\cite{li2022efficientformer}
The EfficientFormer, a family of models, introduces a new dimension-consistent design paradigm for vision transforms, incorporating a simple but efficient latency-driven slimming technique \cite{li2022efficientformer}.
Instead of reducing the number of parameters or computations, EfficientFormer networks are designed to optimize inference speed.
Within this family, the EfficientFormer-L1 is the fastest model, whereas the EfficientFormer-L3 and EfficientFormer-L7 are the largest models, offering better performances.

    \item \textbf{ViT-B/16-DINO\cite{caron2021emerging}:} DINO (DIstillation with NO labels) is a self-supervised learning approach for training vision models, such as ViTs, without the need for labeled data. It relies on a teacher-student framework with a distillation loss and noisy labels to guide the student model. In this work, we consider the ViT-B/16-DINO variant, which corresponds to the DINO approach applied to the ViT-B/16 model using IN-1k.

    \item \textbf{ViT-B/16-SAM\cite{chen2022vision}:} The incorporation of the Sharpness-Aware Minimizer (SAM) into the ViT-B/16 model is an approach that explicitly smooths the loss geometry during training, leading to improved generalization capabilities. By utilizing SAM, the enhanced ViT-B/16 model not only achieved better accuracy and robustness compared to ResNets with similar and larger sizes, but also demonstrated effective training with (momentum) SGD. 
    
    \item \textbf{DeiT-B/16\cite{touvron2021training}:} Standing for Data-efficient image Transformers (DeiT), this method improves the data efficiency of ViTs by employing knowledge distillation, a technique that transfers knowledge from a larger pre-trained teacher model to a smaller student model. DeiT-B/16 is a specific configuration of the DeiT model that adopts the same $16\times16$ patch size, 12 layers, and 768 hidden dimension size as its counterpart in the ViT family.
    
    \item \textbf{DeiT3-B/16\cite{touvron2022deit}:} The DeiT3 method introduces a new training procedure for ViT architectures and is an upgrade of the previous DeiT. Key experiments conducted by the author involve: adopting a binary cross-entropy loss for IN-1k training; comparing simple random cropping to random resize cropping when pre-training in larger datasets such as IN-21k; and training models at lower resolutions to reduce the train-test discrepancy. DeiT3-B/16 refers to a ViT-B/16 model trained using the DeiT3 methodology. Here, we employ two variants of this model: with IN-1k pre-training, or using IN-21k then fine-tuning on IN-1k.

    \item \textbf{CrossViT-B~\cite{chen2021crossvit}:} The Cross-Attention Multi-Scale Vision Transformer (CrossViT) introduces a dual-branch transformer architecture. In one branch, the model processes fine-grained small patches from the image, while the other branch focuses on coarse-grained large patches. This design aims to generate more significant features by incorporating information from different scales. Also, the work proposes a token fusion mechanism with linear complexity, which combines the class token from one branch with the other patches, and vice versa. The architecture is trained based on the approach outlined in DeiT~\cite{touvron2022deit}.
    
    \item \textbf{ConViT~\cite{d2021convit}:} The ConViT architectures tries to mimic the convolutional inductive bias introducing a new attention scheme, the Gated Positional Self-Attention (GPSA). This mechanism forces the attention to initialize following an almost convolutional configuration adding parameters related to attention center and attention locality and then adapting the parameters during the training step. This work shows almost the same accuracy on ImageNet-1k as DeiT~\cite{touvron2022deit} using only 50\% of the dataset, demonstrating the benefits of trying to mimic the inductive bias from CNNs. The architecture is trained based on the approach from DeiT~\cite{touvron2022deit} and the ConViT-B has almost the same number of parameters in comparison to ViT-B.

    \item \textbf{GC ViT-B~\cite{hatamizadeh2022global}:} This variant proposes a method to combine both the standard local context in self-attention with a global context, alternating both blocks to capture fine- and coarse-grained features. The global self-attention makes it possible to query image regions instead of patches (overlapping patches) by applying a convolutional layer. This work shows a greater accuracy on ImageNet-1k compared to ViT-B~\cite{dosovitskiy2020image} with almost the same number of parameters.

    \item \textbf{MViTv2-B~\cite{li2022mvitv2}:} The improved Multiscale Vision Transformers (MViTv2) architecture was proposed to work on both image and video domains. This architecture encodes relative position information in the self-attention and uses a pooling operation after the linear projection on both $Q$, $K$, and $V$ inside the transformer block.

    \item \textbf{CaiT-S24~\cite{touvron2021going}:} This work proposed a method to make deeper ViT possible without saturating the accuracy. The proposed method divides the architecture into two stages, the self-attention and the class-attention, where the first is identical to ViT, except that it has no class token, and the second comprises the patches into the class embedding and extracts more fine-grained patches to the class token, increasing the accuracy and making the training of deeper architectures viable. The CaiT-S24 architecture shows a greater accuracy on ImageNet-1k than ViT-B~\cite{dosovitskiy2020image} with almost half of the parameters. Also, the training schedule was based on DeiT~\cite{touvron2022deit}.

    \item  \textbf{XCiT-M24/16~\cite{ali2021xcit}:} This variant proposes a new self-attention mechanism that decreases the quadratic cost of the original approach. This is achieved by Cross-Covariance Attention, which operates across feature channels rather than tokens, resulting in a linear complexity in the number of tokens. This architecture is more efficient for processing high-resolution images and has a better scalability than the original ViT. The XCiT-M24/16 architecture used here was pre-trained using the DINO~\cite{caron2021emerging} self-supervised approach.

    \item \textbf{BeiT-B/16~\cite{beit}:} The Bidirectional Encoder representation from Image Transformers (BeiT) is a self-supervised approach that proposes masked image modeling to pre-train vision transformers, according to previous findings with the BERT architecture on large language models. It consists of learning to reconstruct image patches by randomly corrupting some original patches, and it can be applied to previous ViT variants. The B/16 variant corresponds to applying the BeiT self-supervised training framework to the ViT-B/16 architecture. 

    \item \textbf{BeiTv2-B/16~\cite{beitv2}:} This variant improves the previous BeiT self-supervised pre-training by using a semantic-rich visual tokenizer, achieved by vector-quantized knowledge distillation. This technique promotes masked image modeling from pixel level to semantic level, outperforming the previous approach on image classification and semantic segmentation tasks. The model used in this work was firstly pre-trained on ImageNet-1k using the BeiT self-supervised training and then fine-tuned on ImageNet-21k using a supervised approach.

    \item \textbf{Swin-B~\cite{liu2021swin}}: Introduces shifted windows between consecutive self-attention layers. This is achieved by a hierarchical/multi-stage architecture, where the input is firstly split into a common patch embedding (stage~1), and then uses patch merging layers as the depth is increased. For instance, the first patch merging layer (stage~2) concatenates the features of each group of $2 \times 2$ neighboring patches from the original patch embedding. The procedure is then repeated for the following stages using similar window merging approaches, decreasing the output resolution and resulting in a hierarchical representation structure. This approach also incurs linear computational complexity concerning image size, allowing the Swin architecture to show a better compatibility with a broad range of vision tasks. The Swin-B model used in our experiments is pre-trained on the ImageNet-21k dataset in a supervised fashion.
   
\end{itemize}

\subsection{Vision Transformer's features-off-the-shelf}

A common approach to employ foundation models for a novel task, especially in data-scarce scenarios, is to remove the classification head, freeze the pre-trained backbone, and then train only a linear classifier over their ``features-off-the-shelf". In CNNs, this can be achieved by applying Global Average Pooling after convolutional layers or considering the output of fully connected layers. The features of the latter approach, however, are known to be highly correlated with the spatial order of the pixels~\cite{cimpoi2014describing}.

In most of the ViT architectures, the output of the penultimate layer, i.e., the class token, or $x^0_L$ (see Eq.~\eqref{eq:classtoken}), is already suited as an image representation without any additional transformation since it is a 1-dimensional embedding vector. Moreover, the relation of this embedding with the spatial order of the pixels is not direct as in CNNs. In some cases, such as for the Swin architecture~\cite{liu2021swin}, a Global Average Pooling operator is applied over the output feature map of the last transformer layer to obtain the image representation instead of using the class token. In any case, our goal is to analyze how these representations behave for texture analysis. In this context, the ViT itself is not fine-tuned, and the base architecture (backbone) is not modified (except for removing the original output head). This allows us to analyze the potential of existing foundation models when directly applied to texture recognition tasks.

\subsection{Linear classifiers}

The image representations obtained with ViT backbones are used to train linear classifiers. We focus on classifiers that can be trained with less data compared to deep learning models, and which have been studied and used for several decades. Their hyperparameters are also not tuned. The following supervised classifiers from the Scikit-learn~\cite{scikit-learn} Python library are considered:
\begin{itemize}
    \item \textbf{KNN}: $k$-Nearest Neighbors~\cite{fix1951discriminatory} using $k=1$;
    \item \textbf{LDA}: Linear Discriminant Analysis~\cite{ripley2007pattern}, with the least-squares solver and automatic shrinkage using the Ledoit--Wolf lemma;
    \item \textbf{SVM}: Support Vector Machine~\cite{cortes1995support}, with a linear kernel and $C=1$. 
\end{itemize}

After fitting each linear classifier, individually, over the features extracted with a ViT backbone on an image training set, we report the average accuracy of the three classifiers on a test set. This approach is employed to minimize the variance caused by the different classification paradigms, e.g., the features of some ViTs may be better coupled with a specific classifier. For instance, in our experiments, while LDA and SVM performed better in general, the KNN classifier surpassed them in some cases. Therefore, we believe that the average results of the three different classifiers should provide a better overall estimate of the quality of the ViT features in different scenarios.

\section{Experiments and Results}

\subsection{Experimental setup}

Following the methodology described above, we evaluate the ViT variants using an Ubuntu server with two Nvidia GeForce GTX 1080ti graphic cards (11 GB of VRAM each), an Intel Core i7-7820X processor, and 64GB of RAM. The scripts are implemented using PyTorch~\cite{Paszke_PyTorch_An_Imperative_2019}. We employ the PyTorch Image Models library~\cite{rw2019timm} (a.k.a.~timm, using Version 0.6.7) to obtain both the model implementation and pre-trained weights since this is a widely used library in the computer vision community. The ViT feature vectors obtained with timm are then employed for the classification step using the supervised classifiers. We evaluate the performance in terms of the average classification accuracy among the three aforementioned linear classifiers (KNN, LDA, and SVM). 

\subsubsection{Baselines for comparison}

Aside from the ViTs, two additional approaches are used in our experiments as a baseline for comparison. Following the developments in the texture analysis field, we consider the classic Local Binary Patterns (LBP)~\cite{ojala2002multiresolution} method, which was the predominant approach in many computer vision applications before the proliferation of deep learning. While most of the hand-engineered baseline results in this paper refer to LBP, we also include some results using Gabor filters~\cite{hoang2005color}, 3-D RGB histograms, and Improved Fisher Vectors (IFV)~\cite{perronnin2010improving}.

An important difference between hand-engineered and deep learning approaches is their computational cost. For instance, the original implementation of the LBP method has a $O(rn)$ time complexity, where $n$ is the number of grayscale pixels in an image and $r$ is the size of the analyzed neighborhood. In our case, $n=150,528$ for a 224x224 RGB image, and $r$ is usually between 8 and 24, which yields around 1,2 to 3,6 million operations (around 0.001 to 0.004 GFLOPs). A common approach to achieving multiresolution grayscale and rotation invariant LBP descriptors, as in~\cite{ojala2002multiresolution}, is to combine different neighborhood sizes (e.g., 8, 16, and 23), and bin sizes for computing the local binary pattern histogram. We will assume that the cost of LBP is $\approx 0.05$ GFLOPs (and $\approx 256$ features, since this number may vary depending on the parameters), i.e., higher than combining 10 neighborhoods of size 23, which is an overestimate to account for its many possible use cases. Nevertheless, this cost consists of only a fraction compared to deep-learning-based models. Hand-engineered techniques may also benefit from current hardware, e.g., the LBP method may reach a $\theta(1)$ time complexity with recent parallel implementations for GPUs~\cite{badanidiyoor2019theta}.

We also consider the ResNet50 architecture~\cite{he2016deep} as a CNN baseline, which is one of the most frequently used convolutional models. This CNN is pre-trained on IN-1k according to the original source. Additionally, we consider another version of ResNet50 that uses knowledge distillation and IN-21k pre-training~\cite{beyer2022knowledge}. In terms of cost, compared to the base versions of most ViTs, ResNet50 has a lower computational budget but higher feature dimensionality (see Table~\ref{tab:taxonomy}). Further on, we address different aspects of the computational cost in our efficiency analysis (see Section~\ref{sec:efficieny}).

\subsubsection{Texture recognition tasks}
\begin{figure*}[!htb]
    \centering
    \subfigure[Outex10.]{
    \includegraphics[width=0.4\linewidth]{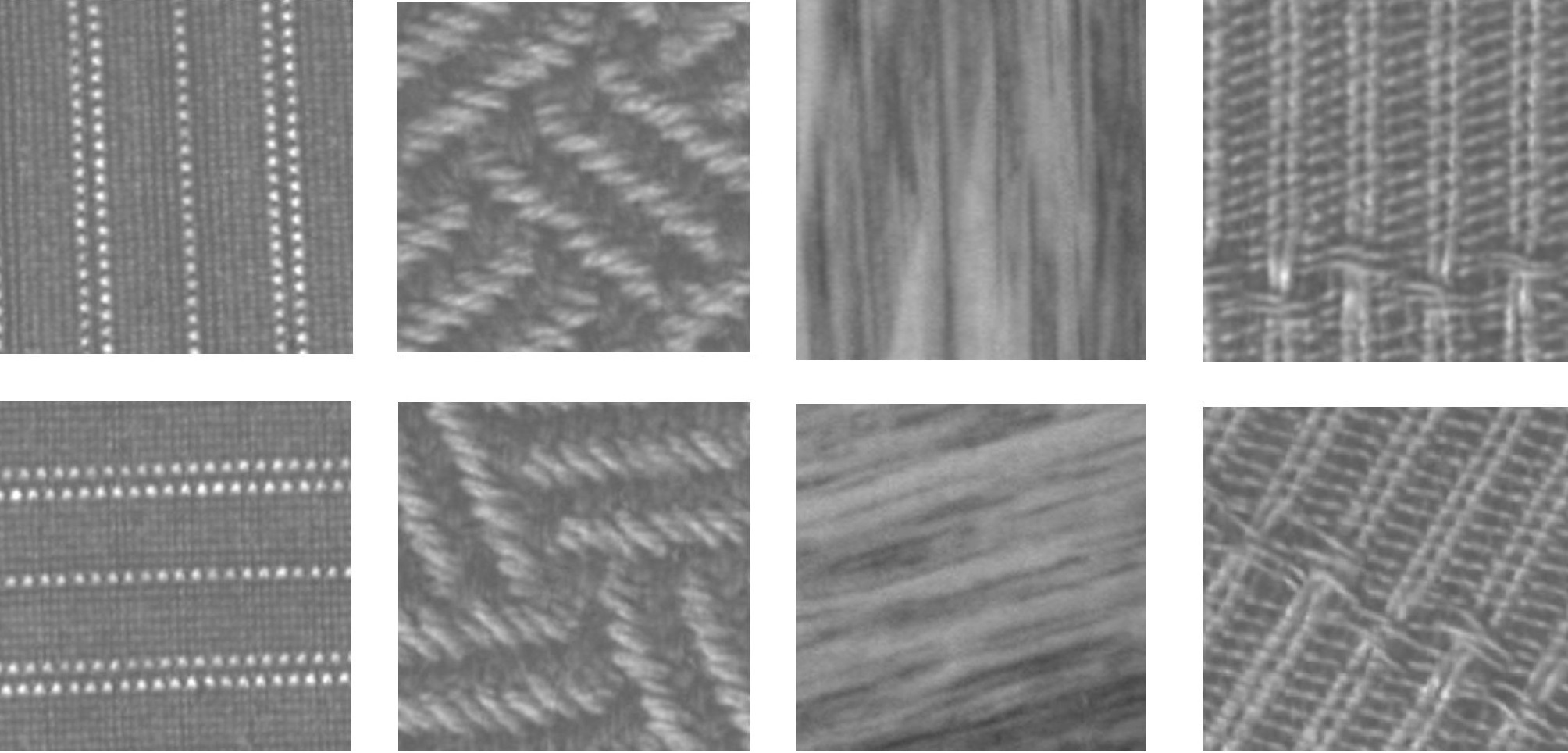}
    } \hspace{1cm}
    \subfigure[Outex11.]{
    \includegraphics[width=0.415\linewidth]{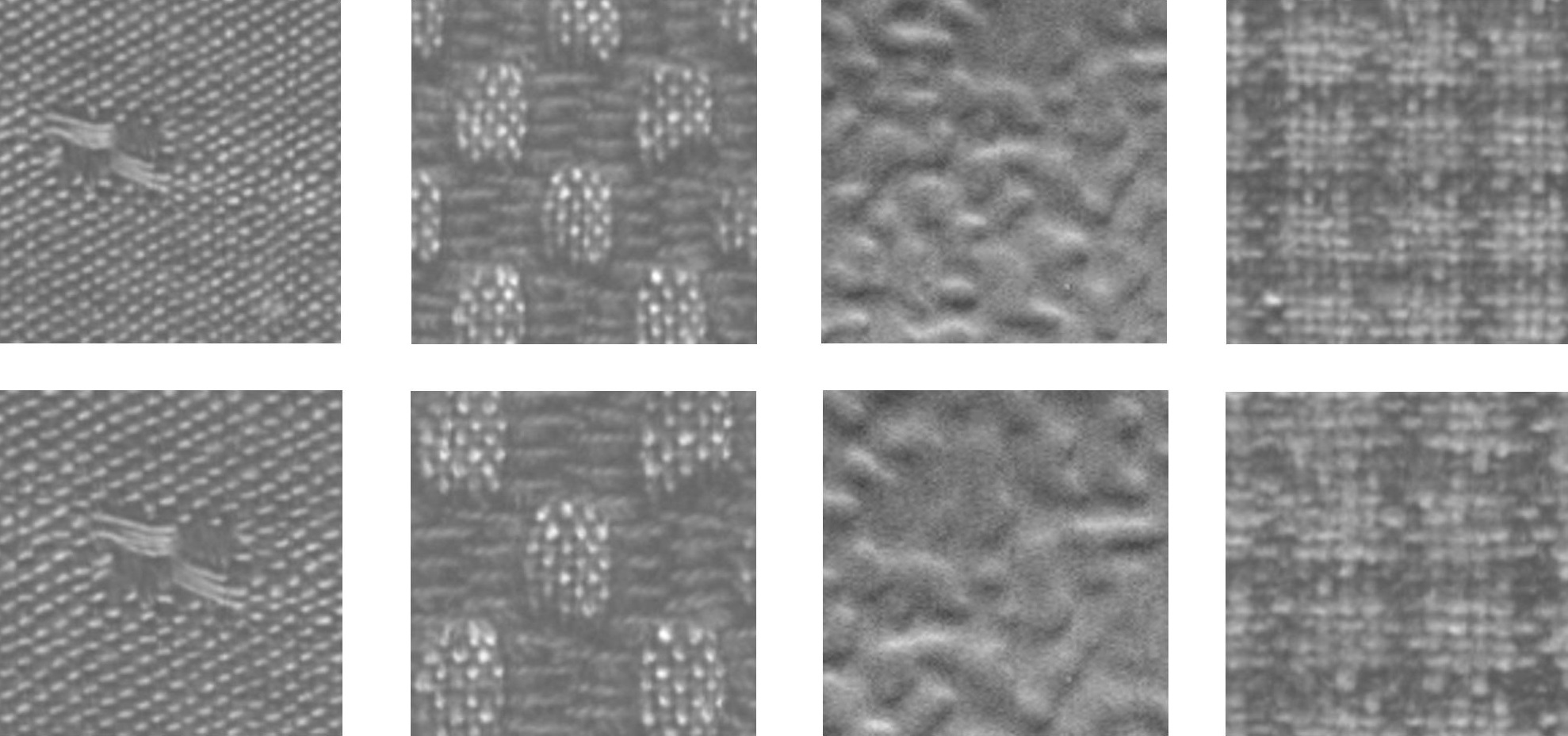}
    }
    \\
    \subfigure[Outex12.]{
    \includegraphics[width=0.415\linewidth]{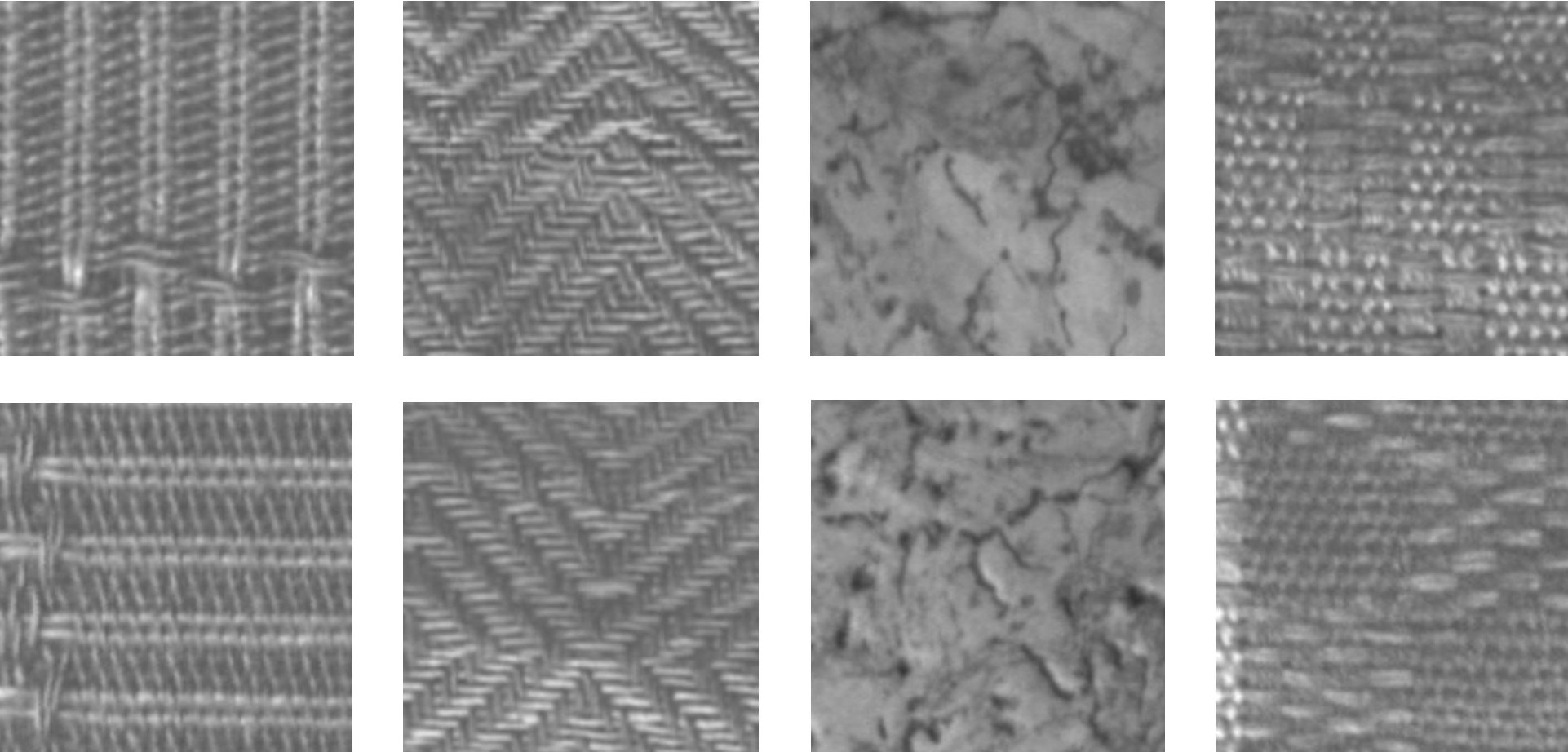}
    }  \hspace{1cm}
    \subfigure[Outex13.]{
    \includegraphics[width=0.41\linewidth]{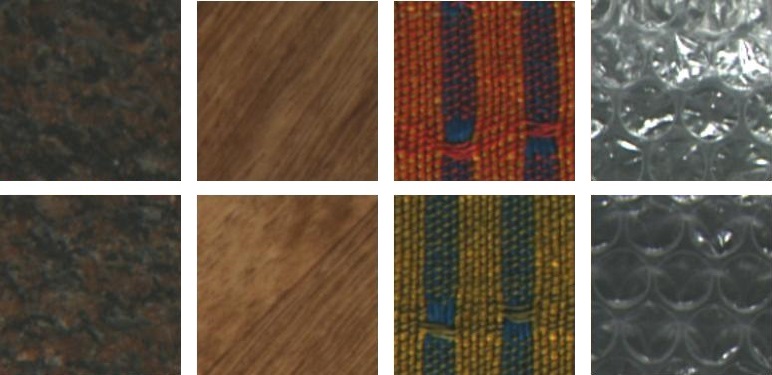}
    } \\
    \subfigure[Outex14.]{
    \includegraphics[width=0.415\linewidth]{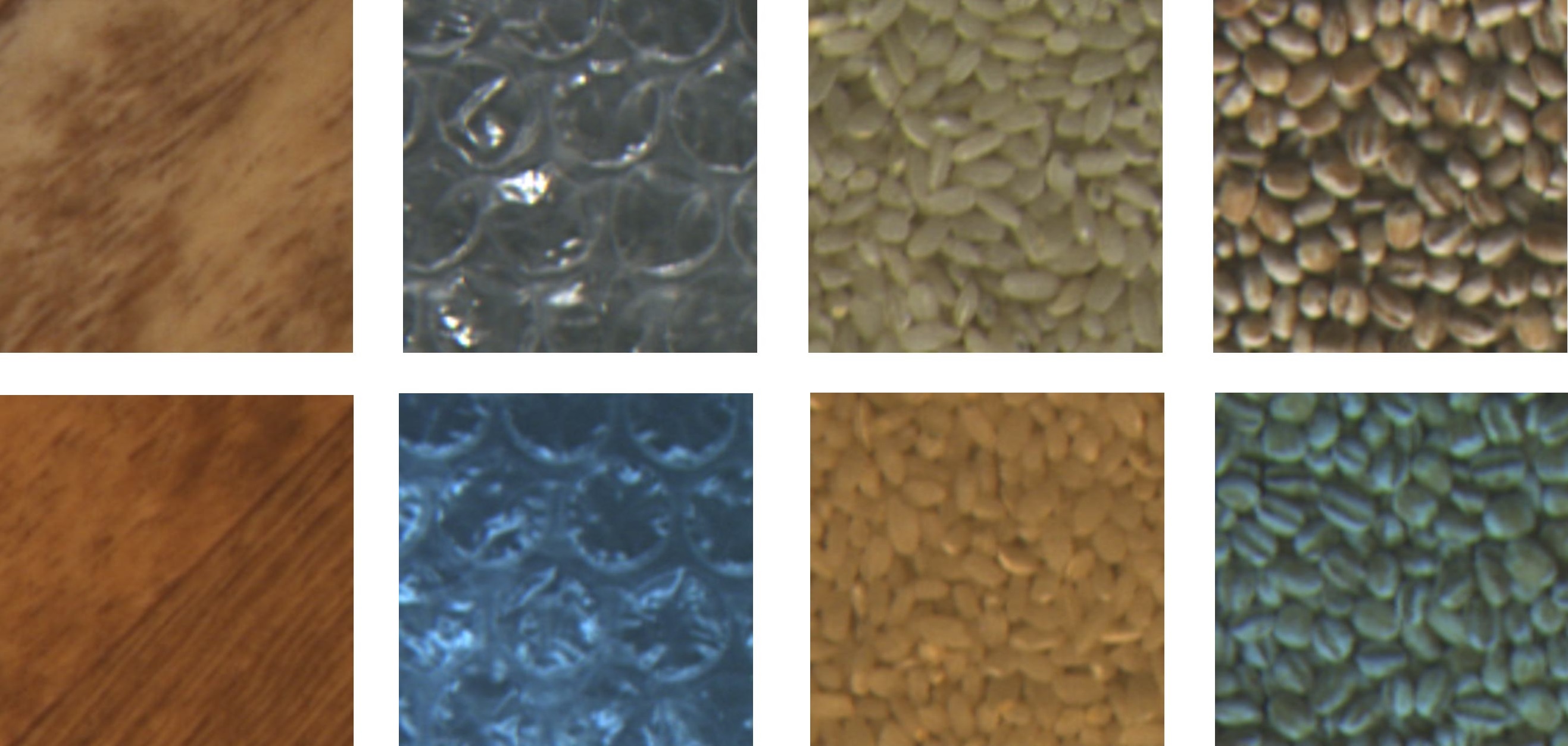}
    } \hspace{1cm} \subfigure[DTD.]{
    \includegraphics[width=0.43\linewidth]{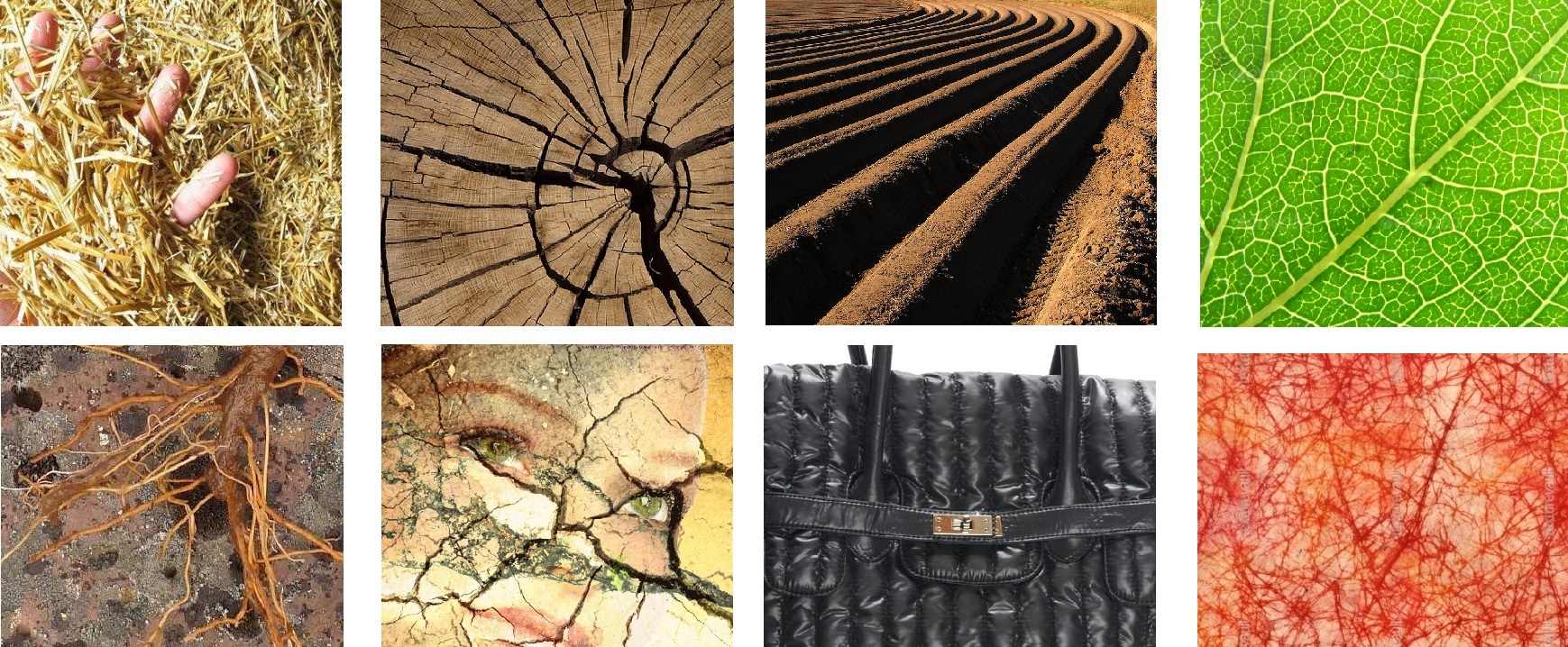}
    } \\
    \subfigure[FMD.]{
    \includegraphics[width=0.42\linewidth]{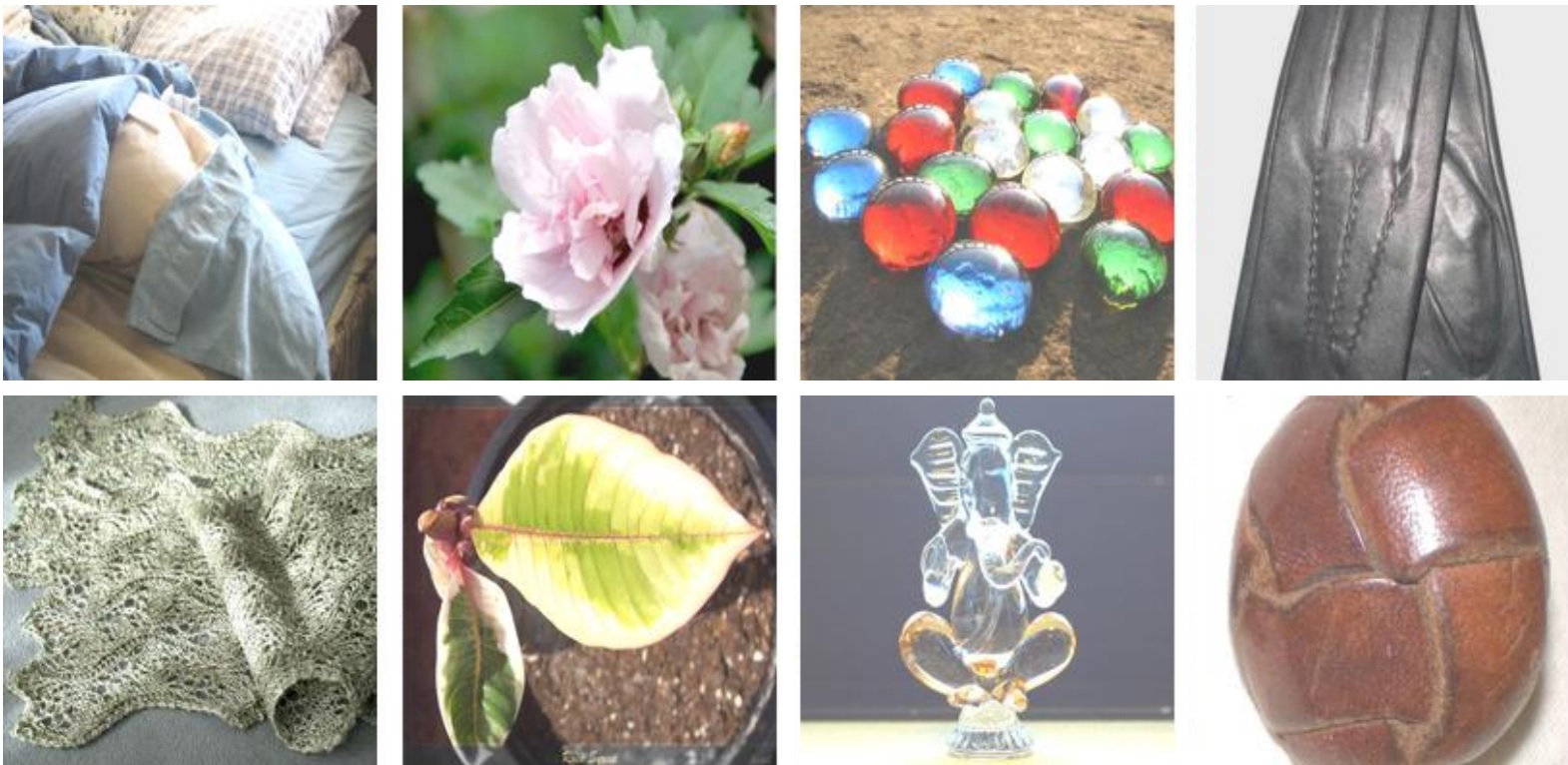}
    } \hspace{1cm} \subfigure[KTH-2-b.]{
    \includegraphics[width=0.43\linewidth]{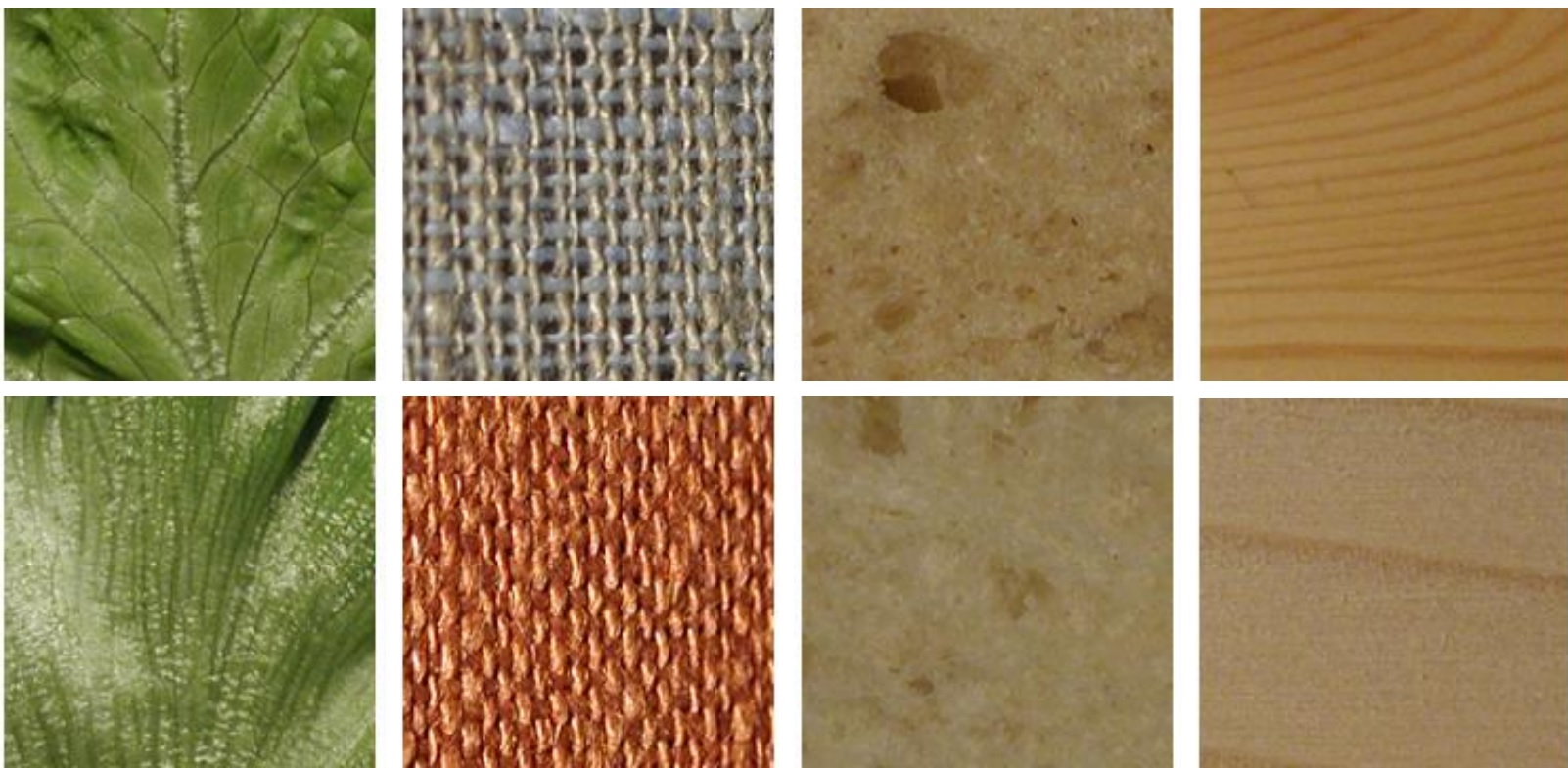}
    }
  \caption{Texture samples from the eight image datasets used in this work. For each dataset, each column represents a different texture class, while each row represent different samples from that class.\label{fig:datasets}}
\end{figure*}

Eight texture datasets are considered in this work in order to analyze a variety of scenarios. They cover several texture recognition tasks such as the classification of materials and texture instances, as well as related properties such as robustness to image transformations. The task difficulty ranges from homogeneous texture images acquired under controlled settings to datasets with a variety of textures taken from the internet. The evaluation policy (training/test splits) also varies among them. We describe each dataset below (Figure~\ref{fig:datasets} shows some samples for each one):
\begin{itemize}
    \item \textbf{Outex10}~\cite{outex}: This dataset consists of 4320 grayscale images belonging to 24 different textures classes, where the train split contains 480 images and the test split contains 3840 images. The same textures are rotated at 9 different angles (0, 5, 10, 15, 30, 45, 60, 75, 90);    
    \item \textbf{Outex11}~\cite{outex}: Consists of 960 grayscale images representing 24 different texture classes, where the train split is composed of 480 images and the test split contains 480 images. This dataset represents textures under different scales; 
    \item \textbf{Outex12}~\cite{outex}: This version is composed of 9120 grayscale images representing the 24 different textures, which is split in two folds, where each fold has the same 480 images in the train split and 4320 test images (two test folds). This dataset represents textures under 9 different rotation angles and different illumination;
    \item \textbf{Outex13}~\cite{outex}: Is composed of 1360 RGB images of 68 texture classes, and evaluates color texture recognition. The samples are split into 680 images for training and 680 images for testing;
    \item \textbf{Outex14}~\cite{outex}: This dataset contains 4080 RGB images corresponding to 68 texture classes, and evaluates color texture recognition under different illumination. The train split contains 680 images, while the test split contains 1360 images;
    \item \textbf{Describable Texture Dataset} \textbf{(DTD)}~\cite{cimpoi2014describing}: Is composed of 5640 images belonging to 47 texture classes, with images taken from the internet with minimal control (textures in-the-wild). Evaluated on the 10 provided splits for training, validation, and test;
    \item \textbf{Flickr Material Dataset (FMD)}~\cite{sharan2010}: Holds 1000 images representing 10 material categories, also obtained from the internet. The validation is done through 10 repetitions of 10-fold cross-validation; 

    \item \textbf{KTH-TIPS2-b}~\cite{caputo2005}: Contains 4752 images of 11 different materials, which are split according to a fixed 4-fold cross-validation. The images have 9 different scales equally spaced logarithmically per sample, 3 camera poses (frontal, 22.5º left and 22.5º right), and 4 illumination conditions (front, from the side at roughly 45º, from the top at roughly 45º, and using ambient lighting).
\end{itemize}

\subsection{Performance comparison}

Our first analysis deals with a general comparison of performance across different texture recognition tasks. These metrics directly reflect the quality of the feature vectors, or image representations, that can be obtained with pre-trained foundation ViTs with open-source code and weights. We divide the texture recognition tasks into two groups: a) theoretical robustness analyses; and b) more complex and realistic tasks. We present the results and corresponding discussion for each group in the following.

\subsubsection{Robustness to geometric transformations and illumination}

We select the Outex10, Outex11 and Outex12 datasets for evaluating the robustness of ViTs. As previously described, these datasets are designed to evaluate the performance of texture recognition models under rotation, scale, and illumination changes. The results for the hand-engineered and CNN baselines, and all ViTs are shown in Table~\ref{tab:results1}.
It represents the average classification accuracy of 
the three linear classifiers after being independently trained and validated according to each dataset cross-validation split. The results are highlighted compared with the baselines and the best results obtained on each dataset. The table is also divided into blocks of rows according to the different approaches for feature extraction: baselines, mobile ViTs, and base ViT models with IN-1k or IN-21k pre-training.

\begin{table}[!htb]
	\centering
	\caption{\label{tab:results1}Average classification accuracy of three
 linear classifiers (KNN, LDA, and SVM) learned over the output features of each ViT backbone. We also list CNN baselines with ResNet50 (with IN-1k and IN-21k pre-training) and the hand-engineered baseline results
  from the Outex (2002)~\cite{outex} paper, based on LBP and Gabor descriptors using the best among a variety of classifiers. \textbf{Bold type} indicates results above the hand-engineered baseline results, and \textcolor{blue}{blue} indicates results above the CNN baseline results (according to pre-training).}
	
	\begin{tabular}{|c|ccc|}
		\hline
		
    	model & Outex10 & Outex11 & Outex12 \\
    	\hline \hline
            hand-engineered baseline & 97.9 & 99.2 & 87.2\\
            CNN baseline (ResNet50)&85.1{\tiny$\pm$1.5}&99.8{\tiny$\pm$0.2}&87.7{\tiny$\pm$1.4}\\
            
\hline

CoaT-Li-Mi&82.6{\tiny$\pm$0.7}&\textbf{99.4{\tiny$\pm$0.2}}&84.4{\tiny$\pm$0.6}\\
CoaT-Mi&85.1{\tiny$\pm$0.9}&\textbf{99.3{\tiny$\pm$0.2}}&\textbf{\textcolor{blue}{88.1{\tiny$\pm$0.9}}}\\ 
MobileViT-S&70.1{\tiny$\pm$0.4}&87.4{\tiny$\pm$3.7}&62.0{\tiny$\pm$3.0}\\ 
MobileViTv2&69.3{\tiny$\pm$0.9}&95.8{\tiny$\pm$2.2}&69.5{\tiny$\pm$0.8}\\ 
EfficientFormer-L1&\textcolor{blue}{86.0{\tiny$\pm$0.1}}&\textbf{99.8{\tiny$\pm$0.3}}&86.4{\tiny$\pm$0.4}\\
\hline

EfficientFormer-L3&\textcolor{blue}{91.3{\tiny$\pm$0.7}}&\textbf{\textcolor{blue}{100{\tiny$\pm$0.0}}}&\textbf{\textcolor{blue}{92.1{\tiny$\pm$0.7}}}\\ 
ViT-B/16-DINO&\textcolor{blue}{93.5{\tiny$\pm$1.3}}&\textbf{\textcolor{blue}{100{\tiny$\pm$0.0}}}&\textbf{\textcolor{blue}{94.0{\tiny$\pm$1.1}}}\\ 
DeiT-B/16&\textcolor{blue}{91.9{\tiny$\pm$0.6}}&\textbf{\textcolor{blue}{99.9{\tiny$\pm$0.1}}}&\textbf{\textcolor{blue}{92.5{\tiny$\pm$0.7}}}\\ 
DeiT3-B/16&\textcolor{blue}{86.1{\tiny$\pm$1.3}}&\textbf{\textcolor{blue}{99.9{\tiny$\pm$0.1}}}&\textbf{\textcolor{blue}{88.0{\tiny$\pm$1.6}}}\\ 
ViT-B/16-SAM&\textcolor{blue}{90.9{\tiny$\pm$1.2}}&\textbf{\textcolor{blue}{100{\tiny$\pm$0.0}}}&\textbf{\textcolor{blue}{91.8{\tiny$\pm$0.8}}}\\ 
CrossViT-B&\textcolor{blue}{87.4{\tiny$\pm$1.5}}&99.2{\tiny$\pm$0.3}&\textbf{\textcolor{blue}{89.1{\tiny$\pm$1.5}}}\\ 
ConViT-B&\textcolor{blue}{86.8{\tiny$\pm$0.2}}&\textbf{\textcolor{blue}{100{\tiny$\pm$0.0}}}&\textbf{\textcolor{blue}{88.8{\tiny$\pm$0.2}}}\\ 
GC ViT-B&\textcolor{blue}{92.5{\tiny$\pm$0.7}}&\textbf{\textcolor{blue}{99.9{\tiny$\pm$0.1}}}&\textbf{\textcolor{blue}{91.8{\tiny$\pm$0.7}}}\\ 
MViTv2-B&\textcolor{blue}{91.2{\tiny$\pm$0.6}}&\textbf{99.7{\tiny$\pm$0.1}}&\textbf{\textcolor{blue}{92.3{\tiny$\pm$0.4}}}\\ 
CaiT-S24&\textcolor{blue}{91.3{\tiny$\pm$0.9}}&\textbf{\textcolor{blue}{99.9{\tiny$\pm$0.2}}}&\textbf{\textcolor{blue}{91.5{\tiny$\pm$0.8}}}\\ 
XCiT-M24/16&\textcolor{blue}{87.7{\tiny$\pm$0.1}}&99.0{\tiny$\pm$0.5}&\textbf{\textcolor{blue}{89.7{\tiny$\pm$0.6}}}\\ 

\hline

CNN baseline (ResNet50 IN-21k)&96.9{\tiny$\pm$1.1}&99.5{\tiny$\pm$0.3}&96.9{\tiny$\pm$1.1}\\

ViT-B/16 (IN-21k)&\textbf{\textcolor{blue}{98.4{\tiny$\pm$0.5}}}&\textbf{99.8{\tiny$\pm$0.2}}&\textbf{\textcolor{blue}{97.2{\tiny$\pm$0.5}}}\\ 
DeiT3-B/16 (IN-21k->1k)&84.1{\tiny$\pm$1.3}&\textbf{99.6{\tiny$\pm$0.0}}&85.7{\tiny$\pm$1.0}\\ 
BeiT-B/16 (IN-21k)&\textbf{\textcolor{blue}{98.4{\tiny$\pm$0.4}}}&\textbf{\textcolor{blue}{100{\tiny$\pm$0.0}}}&\textbf{\textcolor{blue}{97.8{\tiny$\pm$0.3}}}\\ 
BeiTv2-B/16 (IN-21k)&96.5{\tiny$\pm$0.3}&\textbf{\textcolor{blue}{100{\tiny$\pm$0.0}}}&\textbf{\textcolor{blue}{97.1{\tiny$\pm$0.3}}}\\ 
Swin-B (IN-21k)&\textcolor{blue}{97.8{\tiny$\pm$0.6}}&\textbf{\textcolor{blue}{100{\tiny$\pm$0.0}}}&\textbf{\textcolor{blue}{97.0{\tiny$\pm$0.3}}}\\ 

\hline 
\end{tabular}
\end{table}

On the Outex10 dataset, ResNet50 is outperformed by the majority of the ViTs, which was expected seen the limitation of CNNs regarding global patterns and long-range dependencies, which are crucial for analyzing rotated texture images (see examples from this dataset in Figure~\ref{fig:datasets}(a)). On the other hand, only two ViT models have additionally outperformed the hand-engineered baselines, namely ViT-B/16 (IN-21k) and BeiT-B/16 (IN-21k). These results reflect the data-hungry aspects of ViTs, where IN-1k pre-training does not suffice to achieve better rotation robustness than classical methods such as LBP. Additionally, it also shows that ResNet50 is not able to outperform the hand-engineered baseline (LBP) even when using IN-21k pre-training, highlighting a rotation robustness deficit of ImageNet pre-trained models when dealing with controlled images (or homogeneous texture images).

On the Outex11 and Outex12 datasets, the majority of the ViTs outperform both the hand-engineered and CNN baselines, except for the mobile ViTs. Many of the base-sized ViTs exhibit a high robustness (above 99.5\% accuracy) to the changes in texture scales that are present in Outex11. As for Outex12, which contains illumination changes and is a harder task for the baselines, many ViTs also outperform them. In this sense, this result shows that pre-trained ViTs (of base size) using either IN-1k or IN-21k pre-training achieve a strong scale and illumination robustness compared to the baselines. Some small ViTs, such as CoaT-Mi and EfficientFormer-L1, may also be viable in this scenario given their lower computational budgets, even though they do not surpass the baseline performance in all 
cases. Particularly, it is worth to highlight the robustness to illumination changes of ViT models with IN-21k pre-training specifically (except DeiT-B/16), which outperforms the baselines and IN-1k pre-training with a considerable margin.

It is also worth to point to the degraded performance of some models in this first evaluation step, especially MobileViT-S and Deit-B/16. This is related to their higher sensitivity to rotation, scale, and illumination changes in textures. However, the representation obtained from the last transformer layer may not be ideal for this application (homogeneous textures), considering the complexity of these features. In this sense, multi-depth feature engineering and aggregation may be necessary for improving the ViTs in these cases, to benefit from earlier features, as has been done with pre-trained CNNs when ported to texture analysis~\cite{scabini2023radam}.

\subsubsection{Complex and in-the-wild texture recognition}

The second evaluation step focuses on texture recognition datasets representing more challenging scenarios. The datasets may contain variations in rotation, scale, and illumination as the previous datasets, but the classification tasks become harder due to a series of other factors. We consider five datasets: Outex13 (color textures), Outex14 (color textures with illumination changes), DTD (texture attributes in-the-wild), FMD (materials in-the-wild), and KTH-TIPS2-b (materials under several conditions). The results for all ViTs and baselines are shown in Table~\ref{tab:results2}.

\begin{table}[!htb]
	\centering
	\caption{\label{tab:results2}Average classification accuracy considering more challenging texture recognition tasks involving color, different materials, and patterns collected “in the wild”. The hand-engineered results  
 for Outex are from~\cite{outex}, using a 3-D RGB histogram (Outex13) and LBP (Outex14). 
 The hand-engineered results for DTD and FMD, based on the IFV method, are 
 from~\cite{cimpoi2014describing}, and the KTH results using LBP are from ~\cite{caputo2005}. 
 }
	
	\begin{tabular}{|c|ccccc|}
		\hline
		
    	model & Outex13&Outex14&DTD&FMD& KTH-2-b \\
    	\hline \hline
            hand-engineered baseline &  94.7 & 69.0 & 61.2 & 58.2 & 84.0 \\
            CNN baseline (ResNet50)&87.6{\tiny$\pm$2.2}&54.4{\tiny$\pm$0.7}&69.2{\tiny$\pm$2.9}&81.8{\tiny$\pm$3.9}&84.6{\tiny$\pm$1.3}\\ 

\hline

CoaT-Li-Mi&85.8{\tiny$\pm$3.3}&\textcolor{blue}{60.6{\tiny$\pm$2.2}}&\textbf{66.3{\tiny$\pm$4.0}}&\textbf{79.9{\tiny$\pm$5.7}}&\textbf{\textcolor{blue}{88.0{\tiny$\pm$2.6}}}\\
CoaT-Mi&85.0{\tiny$\pm$2.4}&\textcolor{blue}{57.7{\tiny$\pm$2.8}}&\textbf{64.0{\tiny$\pm$2.2}}&\textbf{79.6{\tiny$\pm$4.5}}&\textbf{84.5{\tiny$\pm$2.8}}\\ 
MobileViT-S&85.0{\tiny$\pm$1.8}&23.6{\tiny$\pm$0.8}&19.4{\tiny$\pm$2.6}&27.1{\tiny$\pm$4.1}&56.8{\tiny$\pm$1.7}\\ 
MobileViTv2&74.4{\tiny$\pm$2.6}&31.6{\tiny$\pm$3.9}&18.0{\tiny$\pm$1.2}&25.2{\tiny$\pm$3.2}&58.3{\tiny$\pm$3.1}\\ 
EfficientFormer-L1&\textcolor{blue}{90.0{\tiny$\pm$2.4}}&\textcolor{blue}{64.6{\tiny$\pm$1.5}}&\textbf{\textcolor{blue}{70.0{\tiny$\pm$2.6}}}&\textbf{\textcolor{blue}{83.4{\tiny$\pm$3.7}}}&\textbf{\textcolor{blue}{86.9{\tiny$\pm$1.7}}}\\ 
\hline
EfficientFormer-L3&\textcolor{blue}{89.6{\tiny$\pm$2.1}}&\textcolor{blue}{64.2{\tiny$\pm$1.4}}&\textbf{\textcolor{blue}{70.7{\tiny$\pm$3.1}}}&\textbf{\textcolor{blue}{83.9{\tiny$\pm$4.1}}}&\textbf{\textcolor{blue}{86.2{\tiny$\pm$1.0}}}\\ 

ViT-B/16-DINO&\textcolor{blue}{94.2{\tiny$\pm$1.3}}&\textbf{\textcolor{blue}{78.4{\tiny$\pm$0.9}}}&\textbf{\textcolor{blue}{74.0{\tiny$\pm$2.2}}}&\textbf{\textcolor{blue}{85.2{\tiny$\pm$3.8}}}&\textbf{\textcolor{blue}{88.6{\tiny$\pm$1.0}}}\\ 
DeiT-B/16&\textcolor{blue}{90.8{\tiny$\pm$2.5}}&\textcolor{blue}{65.8{\tiny$\pm$1.2}}&\textbf{67.2{\tiny$\pm$5.4}}&\textbf{79.9{\tiny$\pm$7.6}}&\textbf{\textcolor{blue}{87.4{\tiny$\pm$1.7}}}\\ 
DeiT3-B/16&\textcolor{blue}{87.9{\tiny$\pm$3.1}}&\textbf{\textcolor{blue}{70.0{\tiny$\pm$3.9}}}&\textbf{67.2{\tiny$\pm$4.4}}&\textbf{\textcolor{blue}{83.2{\tiny$\pm$4.4}}}&\textbf{\textcolor{blue}{86.1{\tiny$\pm$3.1}}}\\ 
ViT-B/16-SAM&\textcolor{blue}{92.6{\tiny$\pm$1.8}}&\textcolor{blue}{65.6{\tiny$\pm$1.2}}&\textbf{65.5{\tiny$\pm$4.2}}&\textbf{77.7{\tiny$\pm$4.9}}&80.8{\tiny$\pm$1.5}\\ 
CrossViT-B&\textcolor{blue}{88.8{\tiny$\pm$2.9}}&\textcolor{blue}{63.6{\tiny$\pm$3.1}}&\textbf{63.6{\tiny$\pm$5.4}}&\textbf{77.5{\tiny$\pm$5.4}}&\textbf{\textcolor{blue}{87.9{\tiny$\pm$2.4}}}\\ 
ConViT-B&\textcolor{blue}{89.5{\tiny$\pm$2.5}}&\textcolor{blue}{67.9{\tiny$\pm$2.2}}&\textbf{66.8{\tiny$\pm$5.4}}&\textbf{80.6{\tiny$\pm$9.9}}&\textbf{\textcolor{blue}{86.1{\tiny$\pm$1.8}}}\\ 
GC ViT-B&\textcolor{blue}{89.9{\tiny$\pm$2.8}}&\textcolor{blue}{67.4{\tiny$\pm$3.0}}&\textbf{\textcolor{blue}{70.1{\tiny$\pm$4.6}}}&\textbf{\textcolor{blue}{85.6{\tiny$\pm$5.6}}}&82.1{\tiny$\pm$2.5}\\ 
MViTv2-B&87.4{\tiny$\pm$3.4}&\textcolor{blue}{65.3{\tiny$\pm$1.7}}&\textbf{69.1{\tiny$\pm$4.5}}&\textbf{\textcolor{blue}{83.3{\tiny$\pm$4.5}}}&\textbf{\textcolor{blue}{87.1{\tiny$\pm$1.5}}}\\ 
CaiT-S24&\textcolor{blue}{90.4{\tiny$\pm$3.0}}&\textcolor{blue}{67.1{\tiny$\pm$1.6}}&\textbf{68.3{\tiny$\pm$3.0}}&\textbf{\textcolor{blue}{82.6{\tiny$\pm$5.0}}}&\textbf{\textcolor{blue}{88.1{\tiny$\pm$1.7}}}\\ 
XCiT-M24/16&\textcolor{blue}{88.7{\tiny$\pm$3.4}}&\textcolor{blue}{62.3{\tiny$\pm$1.0}}&\textbf{62.2{\tiny$\pm$5.3}}&\textbf{78.4{\tiny$\pm$5.8}}&\textbf{\textcolor{blue}{86.4{\tiny$\pm$2.8}}}\\ 
\hline

CNN baseline (ResNet50 IN-21k)&90.5{\tiny$\pm$1.3}&66.3{\tiny$\pm$0.5}&74.0{\tiny$\pm$1.9}&86.1{\tiny$\pm$4.1}&84.7{\tiny$\pm$0.4}\\ 

ViT-B/16 (IN-21k)&\textcolor{blue}{92.2{\tiny$\pm$2.4}}&\textbf{\textcolor{blue}{71.0{\tiny$\pm$3.1}}}&\textbf{71.0{\tiny$\pm$3.6}}&\textbf{82.3{\tiny$\pm$6.0}}&\textbf{\textcolor{blue}{86.5{\tiny$\pm$1.3}}}\\ 
DeiT3-B/16 (IN-21k->1k)&88.8{\tiny$\pm$3.2}&\textbf{\textcolor{blue}{71.6{\tiny$\pm$3.2}}}&\textbf{70.1{\tiny$\pm$3.9}}&\textbf{83.9{\tiny$\pm$4.9}}&\textbf{\textcolor{blue}{89.3{\tiny$\pm$1.2}}}\\ 
BeiT-B/16 (IN-21k)&89.0{\tiny$\pm$1.5}&43.3{\tiny$\pm$1.5}&47.9{\tiny$\pm$4.4}&\textbf{60.0{\tiny$\pm$9.3}}&78.6{\tiny$\pm$1.9}\\ 
BeiTv2-B/16 (IN-21k)&\textcolor{blue}{91.6{\tiny$\pm$1.8}}&\textbf{\textcolor{blue}{73.3{\tiny$\pm$1.1}}}&\textbf{\textcolor{blue}{79.1{\tiny$\pm$2.8}}}&\textbf{\textcolor{blue}{90.9{\tiny$\pm$4.6}}}&\textbf{\textcolor{blue}{93.7{\tiny$\pm$1.2}}}\\ 
Swin-B (IN-21k)&89.8{\tiny$\pm$2.5}&\textcolor{blue}{68.6{\tiny$\pm$1.3}}&\textbf{\textcolor{blue}{78.6{\tiny$\pm$2.9}}}&\textbf{\textcolor{blue}{90.5{\tiny$\pm$5.4}}}&\textbf{\textcolor{blue}{88.4{\tiny$\pm$1.0}}}\\ 

 \hline 

	\end{tabular}
\end{table}

The results for the Outex13 and Outex14 datasets, which evaluate the ability to deal with color textures, show that while many ViTs can outperform ResNet50, just a few of them outperform the hand-engineered baselines. For instance, on Outex13 no neural network was able to surpass the results obtained with a 3-D RGB histogram (from~\cite{outex}), which is a considerably simpler method. As for Outex14, which focuses specifically on illumination changes in color textures, the following methods outperform both baselines: ViT-B/16-DINO and DeiT3 on IN-1k; and using IN-21k pre-training with ViT-B/16, DeiT3-B/16, and BeiTv2-B/16. Furthermore, we highlight the results of ViT-B/16-DINO, which achieves the highest performance among the ViTs on these two datasets and represent a considerable improvement on Outex14 (78.4\% versus 69\% from LBP). In conclusion, although the hand-engineered baselines are strong candidates for characterizing color textures, some architectures and improved pre-training approaches for ViTs may be better in some cases.

DTD and FMD are datasets with texture images obtained \textit{in-the-wild} (from the internet). In this case, the texture recognition task is considerably more challenging since models have to deal with a wide variety of scenarios, noise, multiple objects, conditions, backgrounds, etc. In this sense, the models need to deal not only with texture recognition but also with object detection. In these cases, hand-engineered methods struggle in comparison to neural networks as their performance shows. Nevertheless, the hand-engineered baseline can outperform or perform similarly as some of the compared ViTs, especially some small/mobile architectures. In general, most of the base-sized ViTs outperform the hand-engineered baseline, and some of them are also able to outperform ResNet50. We again point to models with IN-21k pre-training, where the gains can be expressive. For these tasks, we highlight EfficientFormer as a strong small-scale architecture, and BeiTv2 and Swin as the best-performing alternatives.

The last dataset analyzed here is KTH-TIPS2-b. Compared to the previous ones, this dataset has different properties that deserve special attention. Firstly, the images are collected in a controlled setting, and textures cover all their area (see Figure~\ref{fig:datasets} (h)). The textures contain several variations such as scale, view angle, and illumination condition. This means that there is no need for background removal, object detection, and similar skills at which deep learning models excel, and the goal is solely to discriminate the target textures. In this sense, this dataset combines various of the transformations analyzed before and is crucial for comparing the capabilities of different approaches. The results show that most methods obtain a similar classification accuracy in the range $84\%$ to $89\%$. However, most of the ViTs outperform the hand-engineered and CNN baselines. This reflects their potential for texture analysis, corroborating that some ViTs can be powerful alternatives for hand-engineered methods and CNNs.

Another important aspect are the differences in architectural design and training schedules among the ViTs. Although most models are considerably similar to the standard ViT in terms of architecture, some differences such as different embedding approaches or self-attention mechanisms may be related to their performance on texture recognition. Moreover, the way the models are pre-trained may also be critical. Firstly, we observe that models with patch embedding tend to perform generally better than convolutional embeddings. This may be related to the fact that most mobile ViTs use convolutional embeddings, which is understandable considering their focus on a lower computational budget. However, the EfficientFormer is situated among the mobile models, but uses patch embedding and generally achieves a considerably higher performance than the other mobile variants, which supports our claim about the superiority of patch embeddings for texture recognition. In terms of the architecture, we note that although the common ViT-B architecture performs well, some variants with different mechanisms such as the Swin transformer (which uses shifted windows) achieve a superior performance.

Considering the pre-training differences among the ViTs, aside from the obvious difference between IN-1k and IN-21k, we note that self-supervised approaches (DINO and BeiT) generally perform better than supervised approaches. However, the DINO approach works considerably better with the basic ViT-B architecture than with the XCiT architecture, suggesting that feature channel attention may not be ideal for textures compared to token attention. BeiT, another self-supervised approach that employs masked image modeling, is among the best methods for homogeneous texture images and basic transformations but struggles with more complex and in-the-wild texture images. Nevertheless, this seems to have improved for BeiTv2, which enhanced masked image modeling from the pixel to the semantic level, making it perform on par with the best methods and also having the best performance for complex and in-the-wild textures.

\subsection{Efficiency analysis}\label{sec:efficieny}


Performance is not the only desirable trait of CV models. Indeed, efficiency is another strong aspect especially when considering low-cost hardware or mobile devices. Therefore, in this section, we discuss the efficiency of the ViTs in terms of feature extraction cost. We removed MobileViT-S and MobileViTv2 from this analysis for better visualization, considering their degraded performance observed in the previous experiments.

\subsubsection{Computational complexity}

\begin{figure}[!htb]
    \centering
     
    \subfigure[Average accuracy on Outex 10, 11, and 12.]{\includegraphics[width=0.3\linewidth]{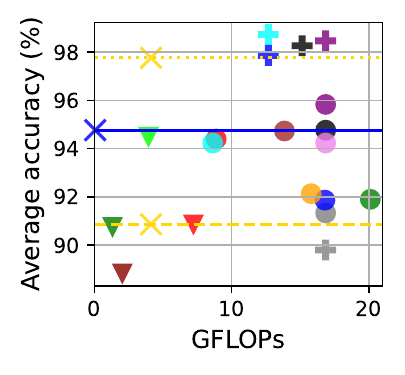} \includegraphics[width=0.2757\linewidth]{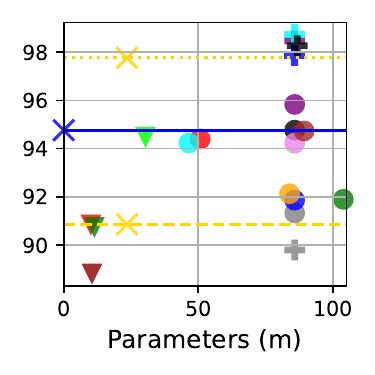} \includegraphics[width=0.2757\linewidth]{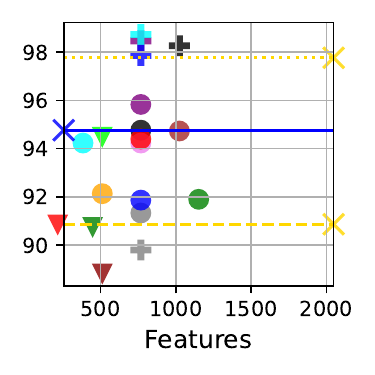}}
    \\
    \includegraphics[width=0.55\linewidth]{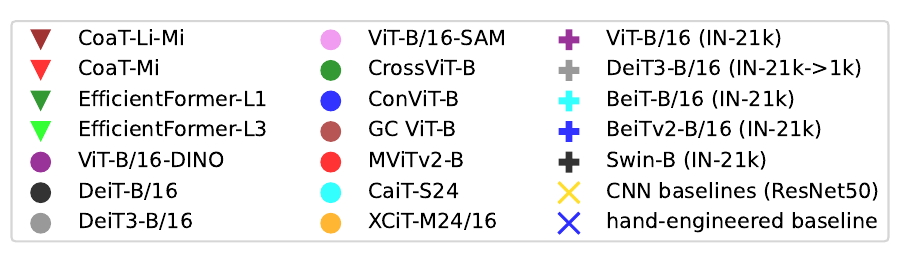}
    \\    
    \subfigure[Average accuracy on Outex13, 14, DTD, FMD, and KTH-2-b.]{\includegraphics[width=0.3\linewidth]{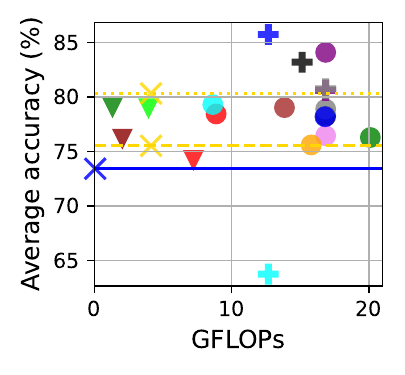} \includegraphics[width=0.2757\linewidth]{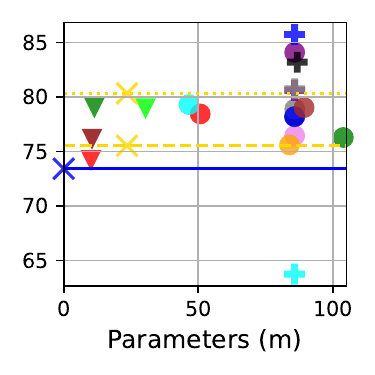} \includegraphics[width=0.2757\linewidth]{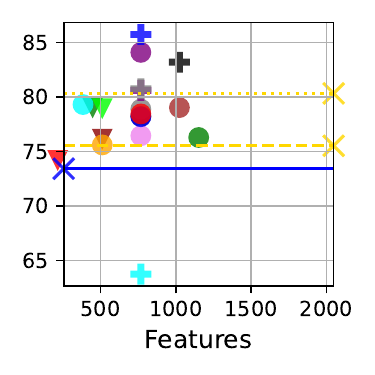}}

  \caption{\label{fig:efficiency}Efficiency analysis of ViT variants compared to hand-engineered and CNN baselines, where accuracy represents the average accuracy over the corresponding datasets and  classifiers considered (KNN, LDA, and SVM). The yellow line with the smaller dots represents ResNet50 with IN-21k pre-training.}
\end{figure}

The computational cost of neural networks or other machine learning models can be estimated using a variety of properties. We consider three measurements that are commonly employed in the deep learning and computer vision literature: the number of floating point operations (FLOPs), the number of parameters, and the number of features (feature vector size). The combination of these measurements provides a consistent estimation of the feature extraction cost for the ViTs under comparison. For instance, the number of FLOPs estimates the processing time, the number of parameters estimates the memory consumption, and the number of features shows the size needed to encode the images (feature dimensionality) as well as the cost of using these features for pattern recognition (classification, regression, etc.). Figure~\ref{fig:efficiency} shows the results of this analysis, where we consider the correlation between the cost measurements and the average performance on the texture recognition tasks.

Figure~\ref{fig:efficiency}(a) considers the average performance on the first three datasets (Outex 10, 11, and 12, see also
Table~\ref{tab:results1}).
These results indicate the correlation between cost and robustness to rotation, scale, and illumination. 
The few ViTs that outperform the baselines have a significantly higher computational cost, except in terms of number of features where ViTs use a smaller image representation than ResNet50. The best alternative is the EfficientFormer-L3, but this ViT variant is outperformed by ResNet50 with IN-21k pre-training. This situation may change with stronger pre-training of mobile or other smaller ViT variants to improve their robustness.

Figure~\ref{fig:efficiency}(b) focuses on datasets representing a more challenging scenario (Outex13, 14, DTD, FMD, and KTH-2-b, see also Table~\ref{tab:results2}). 
In general, ViTs performs better on these more complex tasks, but the highest performance achieved by the bigger variants comes with a considerably higher computational budget than that of ResNet50. 
Nevertheless, the EfficientFormers and Coat-Li-Mi architectures arise as powerful alternatives in this scenario, offering a balance between efficiency and performance compared to the baselines.


\subsubsection{Feature extraction running time}

In practice, the computational budget of deep neural networks may also depend on the quality of implementation, code optimization, hardware, etc. In this sense, we performed an additional experiment to measure the real running time of each ViT when used for feature extraction on one or more images (batches), also referred to as model throughput (images processed per second). We consider running the models using either CPU or GPU processing, and also by varying the batch size (1, 8, or 16). The results for this experiment are shown in Table~\ref{tab:time}. For each case (cell of the table), we run 100 repetitions and compute the average and standard deviation of the processing times.

\begin{table}[!htb]
	\centering
	\caption{\label{tab:time}Throughput (images per second, the higher the better) of models doing feature extraction (average of 100 repetitions) using batches of 224$\times$224 RGB images, performed on a machine with a GTX 1080ti, Intel Core i7-7820X 3.60GHz processor, and 64GB of RAM. Cells in \textcolor{blue}{blue} represent methods with a throughput higher than ResNet50 (CNN baseline) in the respective column.}
	 \resizebox{\linewidth}{!}{
	\begin{tabular}{|c|ccc|ccc|}
		\hline
		
    	 & \multicolumn{3}{c|}{batch size (CPU)} & \multicolumn{3}{c|}{batch size (GPU)}\\
            model & 1 & 8 & 16 & 1 & 8 & 16\\
 \hline

CNN baseline (ResNet50)&44.59\tiny$\pm5.35$&113.12\tiny$\pm9.86$&42.52\tiny$\pm1.23$&158.43\tiny$\pm3.30$&1226.90\tiny$\pm22.17$&2494.15\tiny$\pm44.70$\\ [0.2em]
 \hline 
CoaT-Li-Mi&\textcolor{blue}{48.06\tiny$\pm2.16$}&\textcolor{blue}{141.89\tiny$\pm11.35$}&\textcolor{blue}{58.60\tiny$\pm2.01$}&\textcolor{blue}{208.83\tiny$\pm8.53$}&672.77\tiny$\pm24.71$&1352.99\tiny$\pm24.88$\\ [0.2em]
CoaT-Mi&15.83\tiny$\pm0.66$&47.09\tiny$\pm2.23$&21.84\tiny$\pm0.51$&68.65\tiny$\pm1.22$&180.24\tiny$\pm4.01$&109.42\tiny$\pm0.21$\\ [0.2em]
MobileViT-S&\textcolor{blue}{49.39\tiny$\pm3.10$}&\textcolor{blue}{128.42\tiny$\pm7.43$}&\textcolor{blue}{43.46\tiny$\pm1.73$}&146.92\tiny$\pm3.10$&1193.77\tiny$\pm23.76$&2375.24\tiny$\pm53.52$\\ [0.2em]
MobileViTv2&\textcolor{blue}{52.90\tiny$\pm3.19$}&\textcolor{blue}{148.75\tiny$\pm7.44$}&\textcolor{blue}{49.31\tiny$\pm1.39$}&122.57\tiny$\pm2.11$&974.70\tiny$\pm15.99$&1854.68\tiny$\pm32.93$\\ [0.2em]
EfficientFormer-L1&\textcolor{blue}{67.88\tiny$\pm6.27$}&\textcolor{blue}{218.89\tiny$\pm21.47$}&\textcolor{blue}{83.22\tiny$\pm2.40$}&\textcolor{blue}{169.66\tiny$\pm6.63$}&\textcolor{blue}{1323.82\tiny$\pm30.80$}&\textcolor{blue}{2628.99\tiny$\pm90.08$}\\ [0.2em]
EfficientFormer-L3&33.34\tiny$\pm2.05$&97.24\tiny$\pm4.97$&37.02\tiny$\pm1.21$&105.00\tiny$\pm3.46$&812.24\tiny$\pm14.43$&1646.44\tiny$\pm36.23$\\ [0.2em]
\hline
ViT-B/16&20.38\tiny$\pm1.47$&46.34\tiny$\pm2.26$&21.99\tiny$\pm0.91$&\textcolor{blue}{302.58\tiny$\pm11.11$}&\textcolor{blue}{2315.91\tiny$\pm56.24$}&\textcolor{blue}{4937.86\tiny$\pm111.52$}\\ [0.2em]
DeiT-B/16&20.26\tiny$\pm1.58$&46.10\tiny$\pm2.25$&21.45\tiny$\pm0.77$&\textcolor{blue}{311.27\tiny$\pm13.23$}&\textcolor{blue}{2324.72\tiny$\pm46.65$}&\textcolor{blue}{4979.28\tiny$\pm104.35$}\\ [0.2em]
DeiT3-B/16&19.99\tiny$\pm1.61$&46.16\tiny$\pm2.15$&20.13\tiny$\pm0.62$&\textcolor{blue}{286.87\tiny$\pm17.90$}&\textcolor{blue}{2154.10\tiny$\pm38.75$}&\textcolor{blue}{4586.07\tiny$\pm95.42$}\\ [0.2em]

CrossViT-B&13.11\tiny$\pm1.28$&35.30\tiny$\pm1.41$&15.55\tiny$\pm0.33$&157.73\tiny$\pm5.22$&1173.96\tiny$\pm28.97$&2459.55\tiny$\pm50.21$\\ [0.2em]
ConViT-B&14.71\tiny$\pm0.98$&32.49\tiny$\pm1.08$&14.75\tiny$\pm0.22$&\textcolor{blue}{190.96\tiny$\pm2.42$}&\textcolor{blue}{1448.37\tiny$\pm17.13$}&\textcolor{blue}{3189.61\tiny$\pm278.31$}\\ [0.2em]
GC ViT-B&10.03\tiny$\pm0.73$&25.64\tiny$\pm0.92$&11.44\tiny$\pm0.70$&65.99\tiny$\pm3.16$&236.71\tiny$\pm4.92$&104.65\tiny$\pm0.82$\\ [0.2em]
MViTv2-B&13.30\tiny$\pm0.51$&39.50\tiny$\pm1.34$&17.48\tiny$\pm0.28$&50.38\tiny$\pm0.74$&185.04\tiny$\pm0.42$&106.18\tiny$\pm0.19$\\ [0.2em]
CaiT-S24&17.69\tiny$\pm1.06$&46.14\tiny$\pm1.76$&23.60\tiny$\pm0.51$&108.59\tiny$\pm3.86$&461.23\tiny$\pm11.32$&182.48\tiny$\pm0.56$\\ [0.2em]
XCiT-M24/16&12.71\tiny$\pm0.68$&35.53\tiny$\pm0.92$&18.59\tiny$\pm0.33$&74.12\tiny$\pm1.75$&240.27\tiny$\pm4.87$&133.77\tiny$\pm1.06$\\ [0.2em]

BeiT-B/16&17.99\tiny$\pm1.05$&44.27\tiny$\pm1.74$&20.48\tiny$\pm0.31$&\textcolor{blue}{235.56\tiny$\pm7.69$}&\textcolor{blue}{1750.17\tiny$\pm32.89$}&\textcolor{blue}{3753.43\tiny$\pm52.67$}\\ [0.2em]
BeiTv2-B/16&18.51\tiny$\pm0.33$&45.70\tiny$\pm0.13$&20.53\tiny$\pm0.11$&\textcolor{blue}{233.99\tiny$\pm4.65$}&\textcolor{blue}{1786.92\tiny$\pm19.56$}&\textcolor{blue}{3740.98\tiny$\pm48.88$}\\ [0.2em]
Swin-B&11.41\tiny$\pm0.13$&30.74\tiny$\pm0.07$&16.64\tiny$\pm0.14$&104.62\tiny$\pm1.26$&819.50\tiny$\pm10.15$&292.53\tiny$\pm5.43$\\ [0.2em]

 \hline 
	\end{tabular}
  }
\end{table}

We observe that the ViT throughput, in practice, is more nuanced than their performance or estimated cost. Firstly, it is important to stress that the throughput decreases when increasing the batch size from 8 to 16 on the CPU, due to the fact that this is an 8-core/16-thread processor. The results also show that mobile or low-cost architectures are generally faster than the CNN baseline (ResNet50) when running on the CPU, while the situation changes on the GPU where only the EfficientFormer-L1 outperforms it. On the other hand, considering the larger ViTs, none of them is faster than ResNet50 on the CPU, while some of them can be up to two times faster than the CNN on the GPU. We highlight the ViT-B and DeiT architectures (which are similar but have different codes), which achieve the highest throughputs while running on the GPU.

The nuances in efficiency can be explained by the inherent differences between CNNs and ViTs. Although having a quadratic cost, the self-attention mechanism is highly compatible with parallel processing hardware like GPUs and TPUs, processing images globally in a single pass compared to the more local and sequential processing of CNNs. Additionally, ViTs have more regular memory access patterns, potentially reducing overheads for spatial invariance and benefiting from adaptive computation. While these advantages can lead to shorter processing times for ViTs, it is important to notice that the type and number of CPUs and GPUs, as well as their memory size, can greatly impact the efficiency of both ViTs and CNNs.

\subsection{Attention maps}

\begin{figure}[!htb]
    \centering
     
    \subfigure[Wood texture with cloudy background.]{\includegraphics[width=0.3\linewidth]{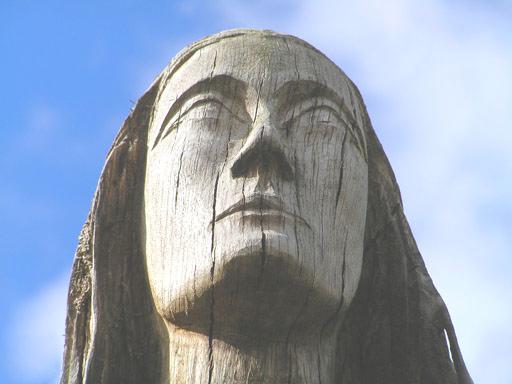}} \subfigure[Glass texture with wooden background.]{\includegraphics[width=0.3\linewidth]{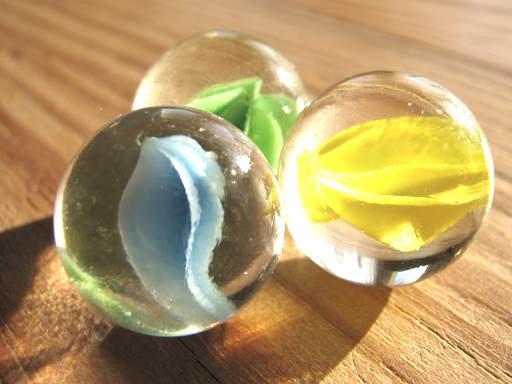}}  \subfigure[Homogeneous wood texture.]{\includegraphics[width=0.3\linewidth]{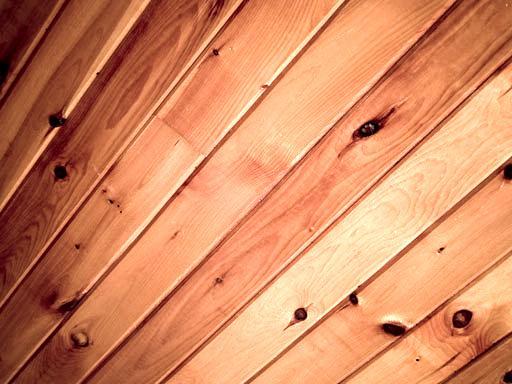}}   
    \\
    \subfigure[ViT-B/16-DINO attention.]{\includegraphics[width=0.3\linewidth]{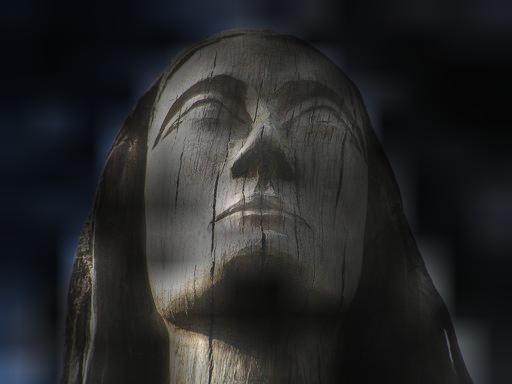} \includegraphics[width=0.3\linewidth]{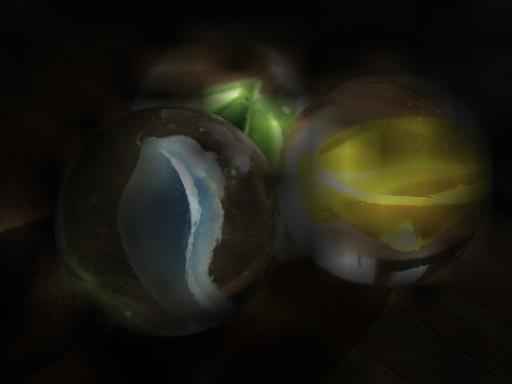} \includegraphics[width=0.3\linewidth]{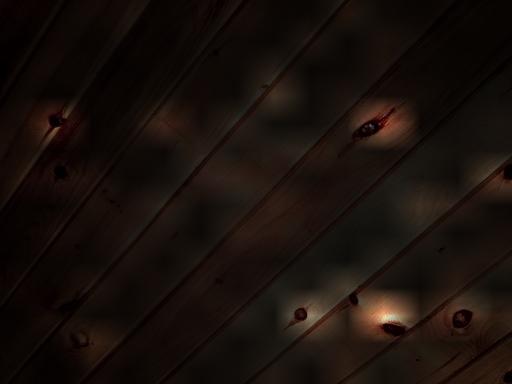}}
    \\
    \subfigure[ViT-B/16 (IN-21k) attention.]{\includegraphics[width=0.3\linewidth]{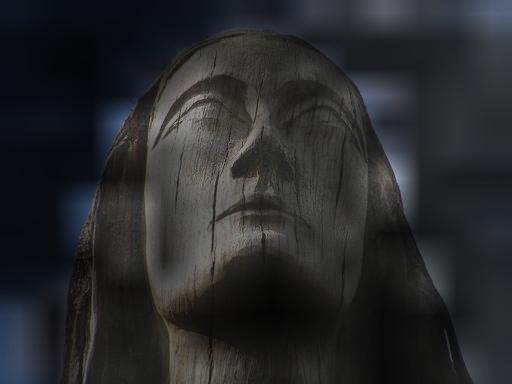} \includegraphics[width=0.3\linewidth]{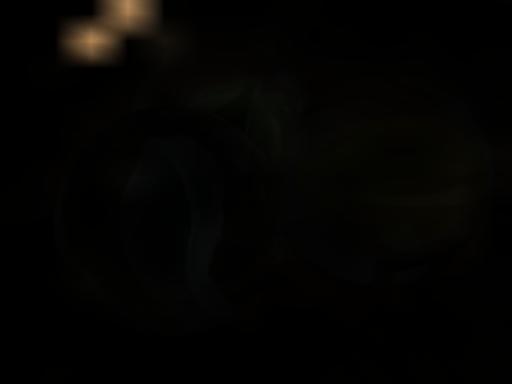}
    \includegraphics[width=0.3\linewidth]{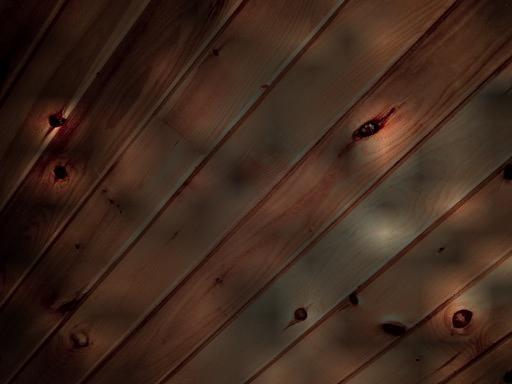}}

  \caption{\label{fig:attention}Visualization of attention maps (at the last layer) of different ViT models (d-e) for texture samples (a-c) from the FMD dataset.}
\end{figure}

To better understand the previously observed performance variation for different ViT pre-trainings, we compute the attention scores for ViT-B/16 using either supervised IN-21k or self-supervised (DINO) IN-1k pre-training. The results are shown in Figure~\ref{fig:attention} for three texture images. The scores are obtained as the average of the output of the softmax operation of a transformer block (layer) in the architecture (see Eq.~\eqref{eq:self-attention}), given an input image. In this sense, for the last layer $l$ we obtain the attention matrix $A_l \in \mathbb{R}^{(n+1) \times (n+1) \times s}$ as the output of the softmax of the self-attention mechanism, where $s$ is the number of attention heads. This matrix is then averaged over all attention heads: 
\begin{equation}
     A_{\mu}(a,b) = \frac{1}{s} \sum_{z=1}^s A_l(a,b,z)\,.
\end{equation}

From $A_{\mu} \in \mathbb{R}^{(n+1) \times (n+1)}$ we first get the scores corresponding to the \textit{class token} (first row excluding its first element), and then reshape it according to the number of patches ($n$), resulting in a $\sqrt{n} \times \sqrt{n}$ matrix. This matrix represents the average attention scores for each token (patch) used on the transformer input. It is then scaled up according to the input image dimensions $w \times h$ (this is the original resolution, not the 224 $\times$ 224 transformation) so that the patches match the original image area. These scores are then used as a mask over the input image for a qualitative analysis of the self-attention operation.

As shown in Figure~\ref{fig:attention}, we selected three different images with similar aspects to better understand the attention mechanism of the different ViTs on textures. The first image contains a wooden statue (labeled as wood), while having a cloudy background. The second image is composed of glass objects (the desired texture), while also having a background with wood texture. The third image does not contain a background and represents only the target texture (wood). While both models can focus on the wooden statue in the first image (with only small differences), the situation is different for the others. The DINO model can effectively focus on the glass texture in the second image, but focuses on the wood defects of the last image which may harm the characterization. On the other hand, the ViT with IN-21k pre-training was not able to focus on the glass texture and the attention seems to collapse on the wooden background, but exhibits a more coherent attention map for the homogeneous wood texture in the last image. This behavior may explain the differences observed between these models on some texture recognition tasks. For instance, IN-21k pre-training achieves a better performance in some cases with homogeneous textures (Outex10 and 12 datasets), while DINO IN-1k pre-training achieves a better performance on the tasks containing textures in-the-wild, i.e., with background, multiple objects, etc (DTD, FMD, and KTH-2-b).

\section{Conclusion}

In this work, we explored several aspects of pre-trained ViTs, a.k.a.\ foundation models, when employed directly for texture analysis by using their class embeddings as image representations. Our analysis shows that ViTs, with their unique architecture and self-attention mechanisms, may provide significant improvements over traditional CNNs and hand-engineered methods on texture recognition tasks. Therefore, the results shed light on the paradigm shift in feature extraction methods in CV. Our experiments compare the features extracted with a variety of ViTs (21 models) for capturing complex texture patterns, their robustness to variations in rotation, scale, and illumination, and the differences between textures filling the whole image or textures in-the-wild (multiple objects, background, etc).

We evaluated the ViT models on eight texture recognition tasks, measured their efficiency, analyzed attention maps, and tested three different linear classifiers as their classification heads. ViTs, through their self-attention mechanism, offer a more global perspective of the image data, unlike the local view provided by CNNs, which is an important aspect of texture analysis. We observe that patch embedding and self-supervised learning are important elements to achieve performant texture discrimination. 
For instance, BeiTv2 and ViT-B/16-DINO demonstrate remarkable performance in general, outperforming other methods, such as ResNet50, with a considerable margin for some tasks. Our results highlight that these models and other ViTs variants can outperform
the hand-engineered baselines and ResNet50 under IN-1k or IN-21k pre-training regimes. These results corroborate the paradigm shift from CNNs to ViTs seen recently in other CV areas.
However, the computation cost of some ViTs may still be a drawback. Some mobile ViT variants may strike a balance between cost and performance, such as the EfficientFormer, as shown by our efficiency analysis. On the other hand, we also show that the throughput of larger ViTs, in practice, can be superior to ResNets on GPUs, which may be related to transformer mechanisms that are more parallelizable and/or better code optimization.

Although showing promising results, our analyses also indicate a need for new techniques and evaluation of more aspects of transformers on textures. Exploring the impacts of different embedding sizes, image resolutions, and model depth will help consolidate their utility in texture analysis. Another aspect is the need for optimized ViT models that balance performance with computational efficiency, making them more accessible for real-world applications. Nevertheless, ViTs emerge as powerful candidates in texture analysis, offering new perspectives and capabilities and corroborating their groundbreaking results in other CV areas. New feature aggregation techniques specifically designed for ViTs and texture may significantly improve the SOTA of texture analysis. Furthermore, as ViTs continue to evolve, they hold the promise of impacting various industries and tasks that rely on texture recognition models.

\section*{Acknowledgements}
L. Scabini acknowledges funding from FAPESP (grants \#2023/10442-2 and \#2024/00530-4). 
L.C. Ribas acknowledges support from FAPESP (grant \#2023/04583-2). 
K.M.C. Zielinski acknowledges support from CAPES (grant \#88887. 631085/ 2021-00) and FAPESP (grant \#2022/03668-1). 
B.~De Baets received funding from the Flemish Government under the "Onderzoeksprogramma Artificiële Intelligentie (AI) Vlaanderen" programme. 
O.M. Bruno acknowledges support from CNPq (Grant \#307897/2018-4) and FAPESP (grants \#2018/22214-6 and \#2021/08325-2).

\bibliographystyle{unsrt}
\bibliography{arxiv.bib}

\end{document}